\documentclass[10pt,twocolumn,letterpaper]{article}

\usepackage{cvpr}              

\usepackage{amssymb}
\usepackage[utf8]{inputenc} 
\usepackage[T1]{fontenc}    
\usepackage{url}            
\usepackage{booktabs}       
\usepackage{amsfonts}       
\usepackage{nicefrac}       
\usepackage{microtype}      
\usepackage{xcolor}         
\usepackage{natbib}
\usepackage{float,wrapfig}
\usepackage{subcaption}
\usepackage{booktabs}
\usepackage{xcolor}
\usepackage{colortbl}
\definecolor{citypink}{RGB}{227, 108, 194}
\usepackage[margin=1in]{geometry}
\usepackage{graphicx}
\usepackage{tikz}
\usetikzlibrary{arrows.meta,shadows,shapes.geometric}
\usepackage{xcolor}
\usepackage{listings}

\usepackage[most]{tcolorbox}
\usepackage{multirow}
\usepackage{array}
\usepackage{longtable}
\usepackage{multirow}
\usepackage{xcolor}
\usepackage{colortbl}
\usepackage{adjustbox}
\usepackage{pifont}
\usepackage{makecell}
\usepackage{array}
\usepackage{tikz}
\usepackage{xspace}
\usepackage{stfloats}
\usepackage[linesnumbered,ruled,vlined]{algorithm2e}
\usepackage{minted}
\definecolor{cvprblue}{rgb}{0.21,0.49,0.74}
\usepackage[pagebackref,breaklinks,colorlinks,allcolors=cvprblue]{hyperref}

\def\paperID{} 
\def\confName{CVPR}
\def\confYear{2026}

\title{OmniLottie: Generating Vector Animations via Parameterized Lottie Tokens}

\newcommand{\ours}{OmniLottie}
\newcommand{\ourdataset}{MMLottie-2M}
\newcommand{\ourbenchmark}{MMLottieBench}

\newtcbox{\highlighttext}[3]{%
  arc=3.5pt, %
  colback=#2!80, %
  boxrule=0pt, %
  left=0.3pt, right=0pt, top=-1.5pt, bottom=-1.5pt,
  nobeforeafter,
  tcbox raise base,
  enhanced,
  fontupper=\textcolor{#1}{#3}%
}

\newtcbox{\highlightgrad}[4]{%
  on line,
  arc=3.5pt, %
  colframe=#3!80!black, %
  boxrule=0pt, %
  left=0.5pt, right=0pt, top=0pt, bottom=0pt,
  nobeforeafter,
  tcbox raise base,
  enhanced,
  valign=center,
  interior style={
    top color=#3!100,
    bottom color=#4!100,
    middle color=#3!50!#4!50,
  },
  fontupper=\textcolor{#1}{\textbf{#2}}%
}
\definecolor{myorange}{RGB}{233,113,50}
\definecolor{myblue}{RGB}{56,162,227}
\definecolor{mygreen}{RGB}{152,204,102}
\definecolor{lottiedata}{RGB}{223,236,247}
\definecolor{svgdata}{RGB}{251,235,226}
\definecolor{dataproc}{RGB}{255,230,230}
\definecolor{annotation}{RGB}{255,255,217}

\newcommand{\lottiejson}{%
  \raisebox{-0.2em}{\includegraphics[width=0.045\linewidth]{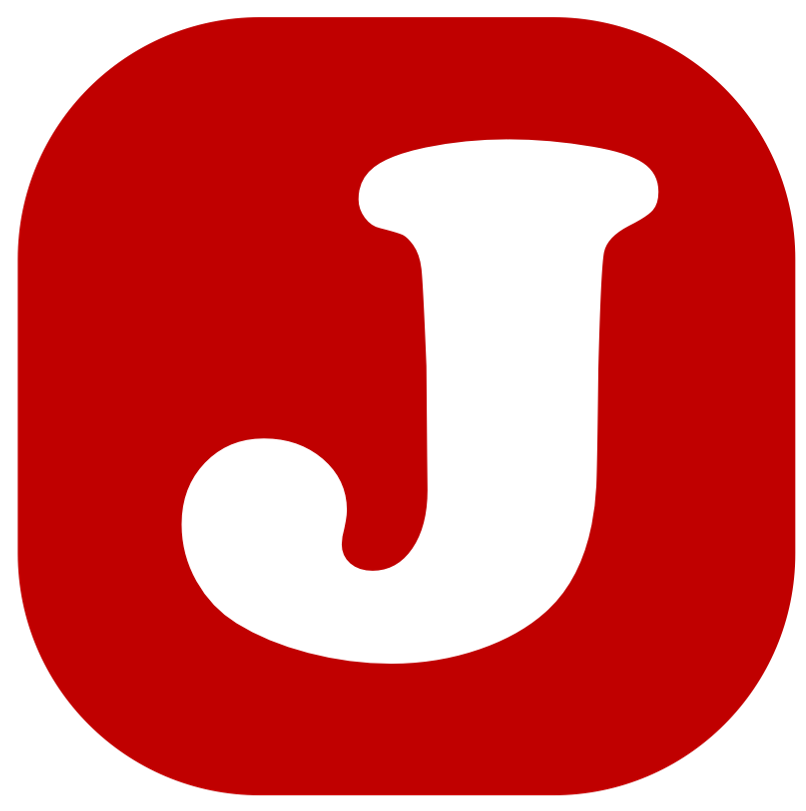}}%
}
\newcommand{\tokenizer}{%
  \raisebox{-0.2em}{\includegraphics[width=0.045\linewidth]{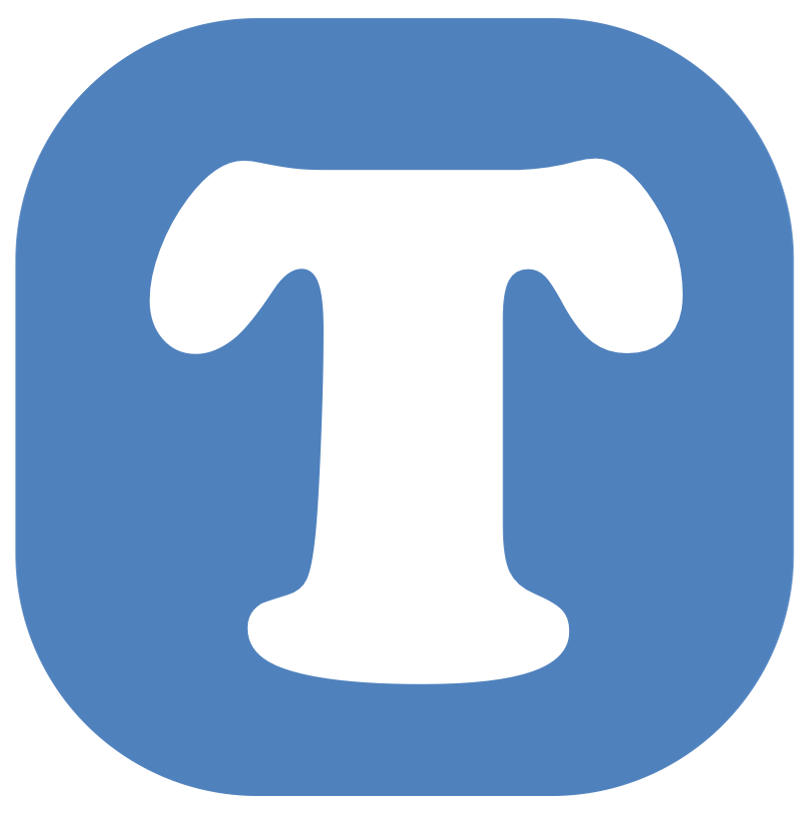}}%
}
\newcommand{\qwen}{%
  \raisebox{-0.2em}{\includegraphics[width=0.045\linewidth]{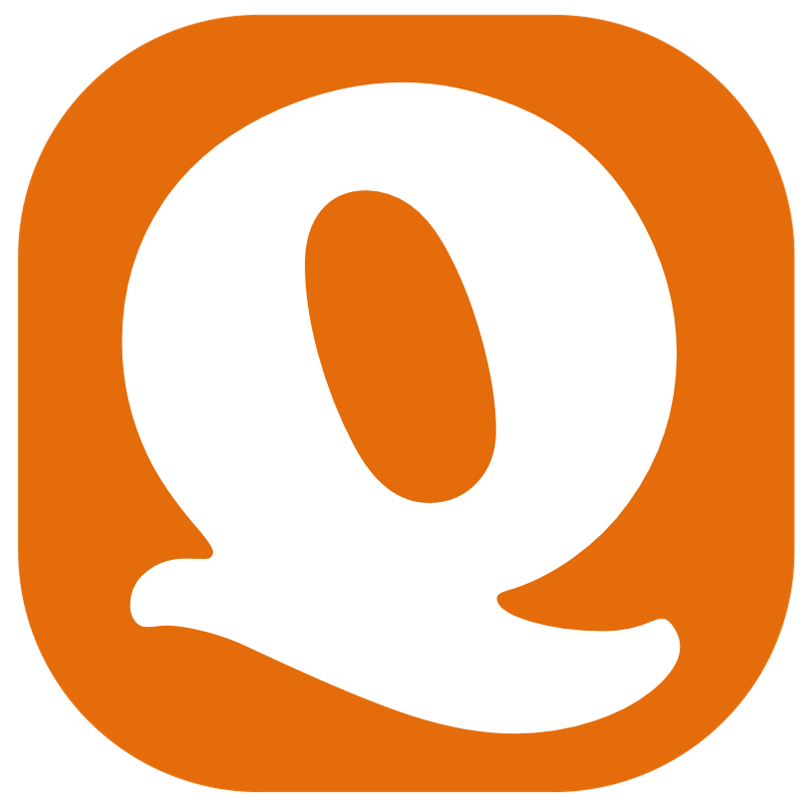}}%
}
\newcommand{\svg}{%
  \raisebox{-0.2em}{\includegraphics[width=0.045\linewidth]{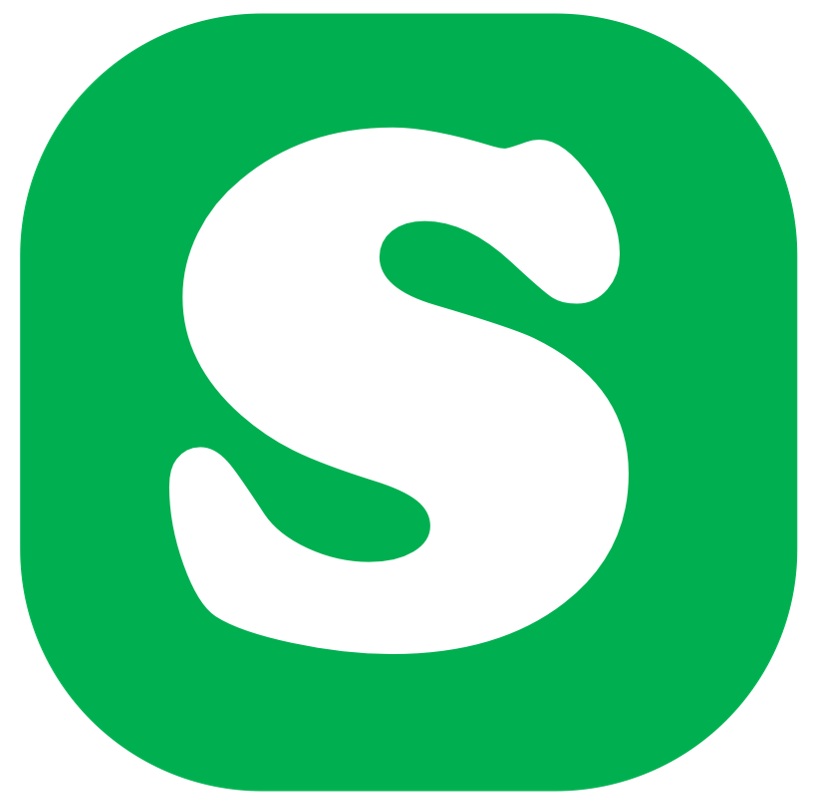}}%
}
\newcommand{\lottie}{%
  \raisebox{-0.2em}{\includegraphics[width=0.045\linewidth]{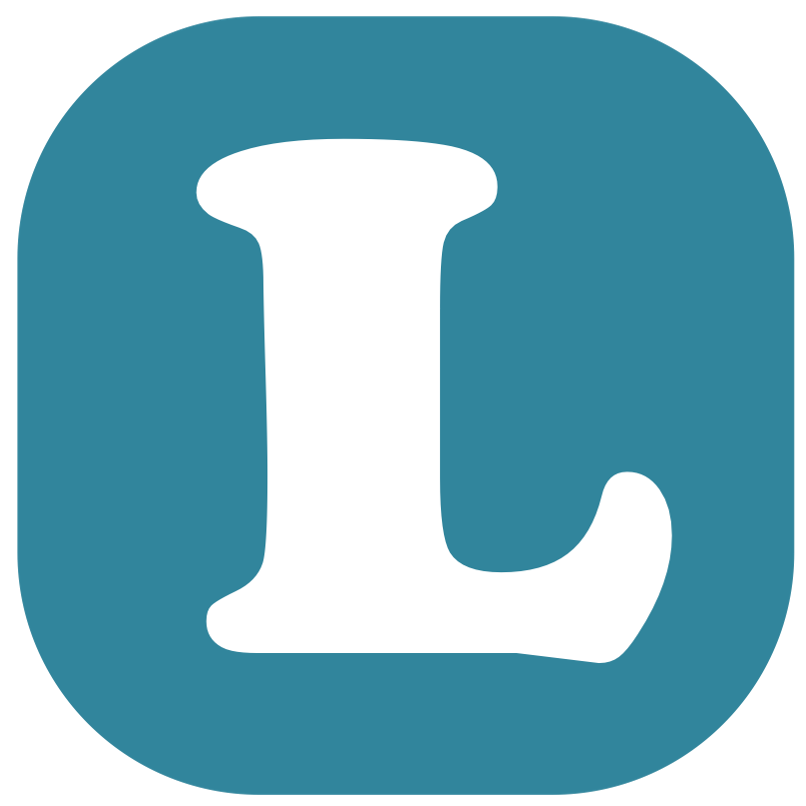}}%
}

\author {
    Yiying Yang\textsuperscript{\rm 1,2}
    \quad
    Wei Cheng\textsuperscript{\rm 2~\dag}
    \quad
    Sijin Chen\textsuperscript{\rm 3}
    \quad 
    Honghao Fu\textsuperscript{\rm 2,4} \\
    Xianfang Zeng\textsuperscript{\rm 2} 
    \quad
    Yujun Cai\textsuperscript{\rm 4}
    \quad
    Gang Yu\textsuperscript{\rm 2 \ddag}
    \quad
    Xingjun Ma\textsuperscript{\rm 1 \ddag} \\ [0.4em]
    $^{1}$ Fudan University
    \quad
    $^{2}$ StepFun 
    \quad
    $^{3}$ HKU MMLab
    \quad
    $^{4}$ University of Queensland
    \\[0.8em]
    \textcolor{citypink}{\normalsize
    \raisebox{-0.2\height}{\includegraphics[height=0.5cm]{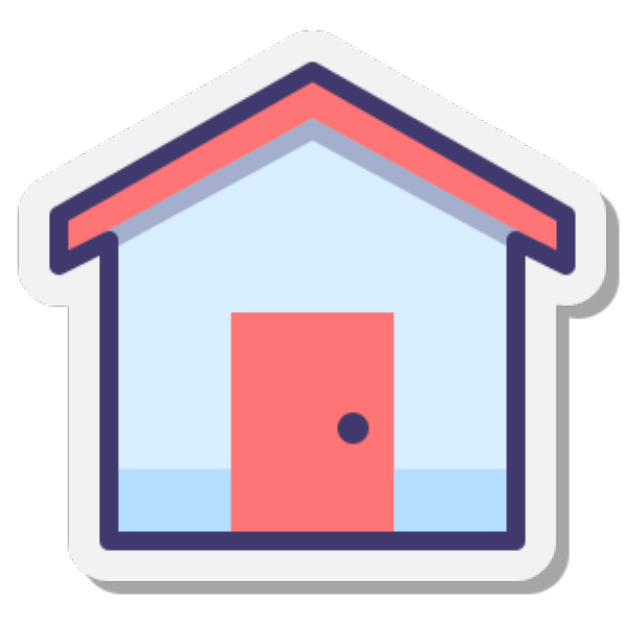}}~{\href{https://openvglab.github.io/OmniLottie/}{\textbf{Project Page}}}
    \quad
    \raisebox{-0.2\height}{\includegraphics[height=0.5cm]{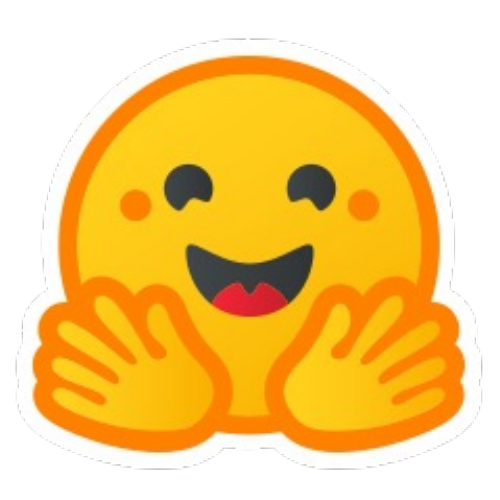}}~{\href{https://huggingface.co/datasets/OmniLottie/MMLottie-2M}{\textbf{\ourdataset}}}
    \quad
    \raisebox{-0.2\height}{\includegraphics[height=0.5cm]{files/huggingface_logo.pdf}}~{\href{https://huggingface.co/datasets/OmniLottie/MMLottieBench}{\textbf{\ourbenchmark}}}
    \quad
    \raisebox{-0.2\height}{\includegraphics[height=0.5cm]{files/huggingface_logo.pdf}}~{\href{https://huggingface.co/OmniLottie}{\textbf{Weights}}}
    \quad
    \raisebox{-0.2\height}{\includegraphics[height=0.5cm]{files/huggingface_logo.pdf}}~{\href{https://huggingface.co/spaces/OmniLottie/OmniLottie}{\textbf{Demo}}}
    \quad
    \raisebox{-0.2\height}{\includegraphics[height=0.5cm]{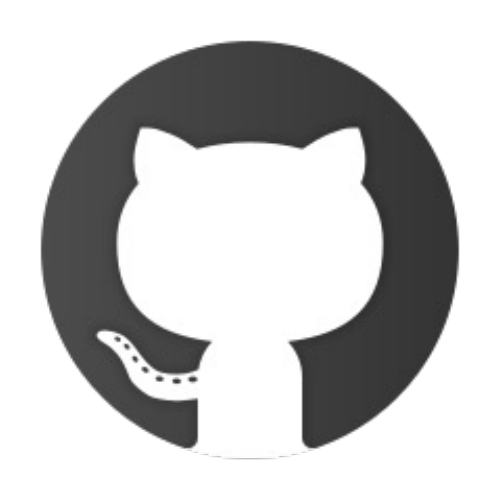}}~{\href{https://github.com/OpenVGLab/OmniLottie}{\textbf{Code}}}
    }
}

\begin{document}

 \twocolumn[{
   \renewcommand\twocolumn[1][]{#1}
   \maketitle
   \begin{center}
     \vspace{-1ex}
     \includegraphics[width=\linewidth]{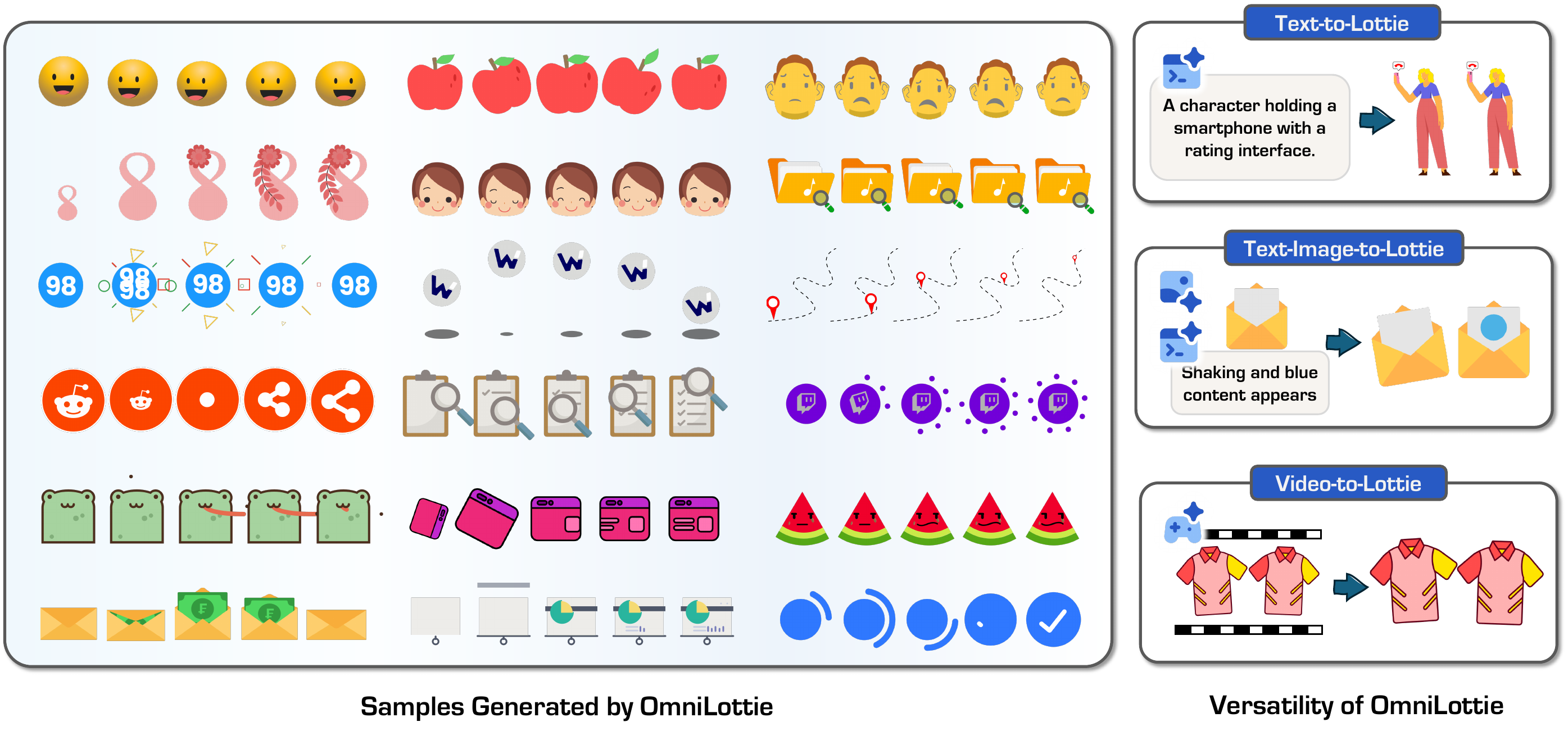}
     \captionof{figure}{\small \textbf{\ours~ is a Versatile Auto-regressive Generative Model for High Quality Lottie Animations}.
     With user inputs of interleaved multi-modal instructions, \ours~supports tasks including text-to-Lottie, text-image-to-Lottie and video-to-Lottie generation.
     This broad capability makes it a powerful and flexible solution for a wide range of creative and design-oriented tasks.}
     \label{fig:teaser}
   \end{center}
 }]

\let\thefootnote\relax\footnotetext{\dag~Project lead. \ddag~Corresponding authors. }

\begin{abstract}

\textbf{\ours} is a versatile framework that generates high-quality vector animations from multi-modal instructions.
For flexible motion and visual content control, we focus on Lottie, a light-weight JSON formatting for both shapes and animation behaviors representation.
However, the raw Lottie JSON files contain extensive invariant structural metadata and formatting tokens, posing significant challenges for learning vector animation generation.
Therefore, we introduce a well-designed Lottie tokenizer that transforms JSON files into structured sequences of commands and parameters representing shapes, animation functions and control parameters. 
Such tokenizer enables us to build \ours~upon pretrained vision–language models to follow multi-modal interleaved instructions and generate high-quality vector animations.
To further advance research in vector animation generation, we curate \textbf{MMLottie-2M}, a large-scale dataset of professionally designed vector animations paired with textual and visual annotations. 
With extensive experiments, we validate that \ours~can produce vivid and semantically aligned vector animations that adhere closely to multi-modal human instructions.

\end{abstract}

\section{Introduction}\label{sec:intro}

Vector animation is a type of computer animation created using vector graphics rather than raster videos~\cite{BLI,AniClipart,WW}.
Compared to raster videos, vector animations provide a lightweight, editing-friendly, and resolution-independent representation that facilitates intuitive manipulation of appearance, stylistic effects, and motion attributes.
Therefore, vector animation is widely adopted across modern design workflows.
Developing autonomous systems capable of generating vector animations from multi-modal instructions has the potential to significantly lower the entry barrier for creation and greatly accelerate professional design pipelines.

Prior approaches for vector animation generation mainly focus on applying motion priors derived from either a reference GIF~\cite{WW} or a text-to-video generator~\cite{AniClipart,BLI,makeavideo,modelscope,tuneavideo} to manually designed static vector graphics. While considering a larger scope of computer animations, the diffusion-based video generation models~\cite{wan2025,kong2024hunyuanvideo,huang2025step,magi1,zhang2025packing,mikudance} have also demonstrated remarkably capability in generating rasterized image animation from texts and images.
Though the above mentioned models are able to generate visually striking animations, their outputs inherently lack editability, cross-platform compatibility, and resolution scalability provided by vector formats.

In this work, we focus on the \textbf{Lottie} representation, a widely adopted vector animation format that stores all the shape, effect, and motion parameters within a single JSON file.
Lottie is a complex \textbf{hierarchical representation} with layers and inheritance to integrate visual elements and motion effects. 
Compared to another widely adopted vector graphics representation, \textit{i.e.} SVG stacks rects, circles, and paths for shapes and embeds CSS codes for animation. 
The superior cross-platform compatibility of Lottie makes it more popular than SVG.
With the rapid advancement of Large Language Models (LLMs), we notice some existing models~\cite{Qwen2.5VL,GPT-5,Gemini} are already able to generate Lottie animations through producing JSON text files, but suffer from unsatisfactory generation success rates (Tab.~\ref{tab:quantitative_sota}) because of the strict formatting of Lottie's nested JSON representations and instruction-following performance (\cref{fig:qualitative_t2l,fig:qualitative_i2l,fig:qualitative_v2l}).

We argue that directly generating raw Lottie JSONs is inefficient: formatting tokens and metadata take up large portions of the tokens rather than meaningful geometric or motion cues.
In order to improve the robustness, efficiency and reliability of vector animation generation, we propose a \textbf{Lottie tokenizer} that converts Lottie files into compact sequences of shape, effect, and animation commands with their associated parameters.

The proposed serialization of Lottie tokens enables us to train auto-regressive models end-to-end on large scale datasets using the best practices from VLM training. To this end, we construct \textbf{\ourdataset}, a comprehensive dataset of vector animations annotated with text descriptions, keyframe images, and rendered videos, to facilitate researches on Text-to-, Text-Image-to-, and Video-to-Lottie generation under a unified framework.
Building on \ourdataset, we further establish standardized evaluation protocols tailored to multi-modal Lottie generation.
Our experiments demonstrate that \ours~significantly outperforms strong baselines in both visual fidelity and semantic alignment with multi-modal inputs, delivering high-quality vector animations across diverse prompts and modalities.

In summary, our contributions are as follows:

\begin{itemize}
\item We introduce \textbf{\ours}, the first end-to-end framework capable of generating vector animations directly from multi-modal instructions.
 \item We curate \textbf{\ourdataset}, a large-scale multi-modal dataset of two million Lottie animations with paired text descriptions, reference images, and
  video demonstrations, which has been publicly released to facilitate future research.
\item We propose a novel \textbf{Lottie tokenizer} that converts raw JSON into concise command sequences, improving both training efficiency and generation quality.
\item Extensive evaluations show that \ours~produces high-quality vector animations and achieves SOTA performance in both visual fidelity and semantic alignment.
\end{itemize}

\begin{figure*}[t]
    \centering
    \includegraphics[width=\linewidth]{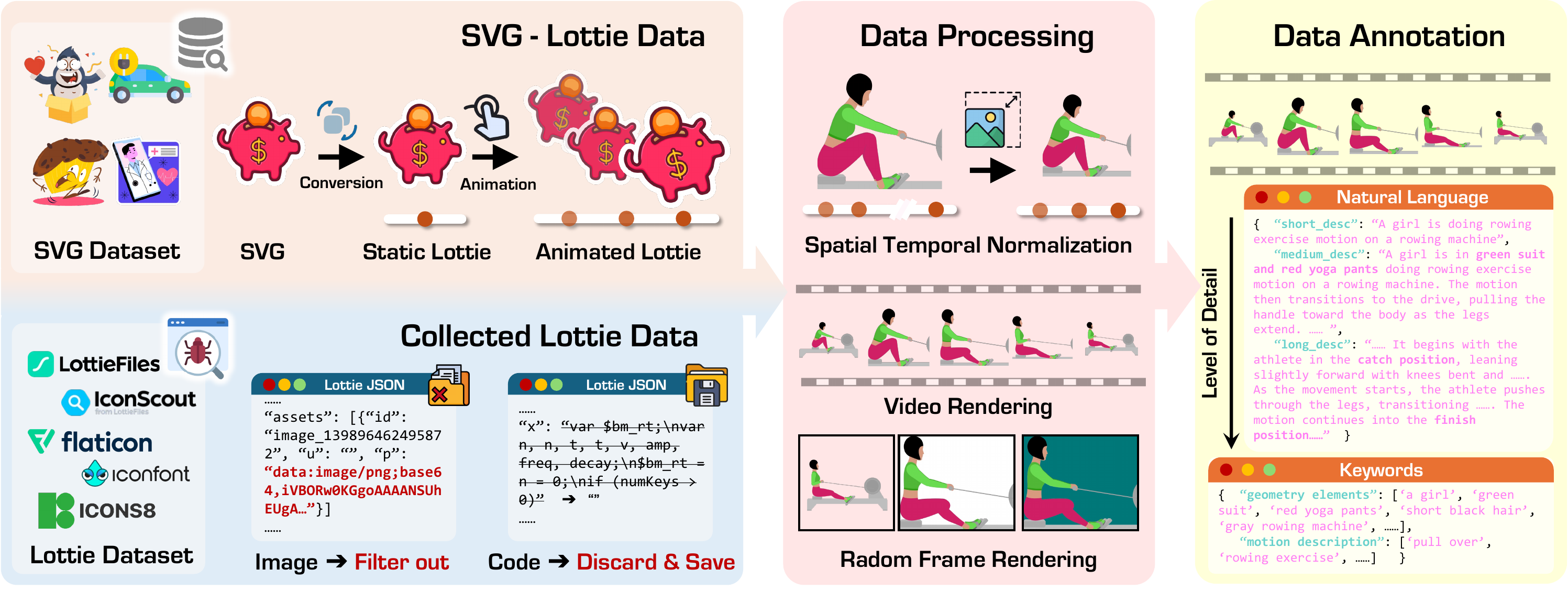}
    \vspace{-4ex}
    \caption{\small \textbf{Overview of the Vector Animation Data Construction Pipeline.}
\highlightgrad{black}{\textbf{Data Collection \& Conversion}}{svgdata}{lottiedata}~: We convert SVG assets into static Lottie files and apply randomized animation effects to generate effect-tagged animated Lotties. In parallel, we gather professionally created Lottie animations from five online platforms and perform thorough filtering and cleaning.
\highlighttext{black}{dataproc}{\textbf{Data Processing}}~: Each animation undergoes spatio-temporal normalization, followed by video rendering and random keyframe extraction.
\highlighttext{black}{annotation}{\textbf{Data Annotation}}~: Finally, we provide multi-granularity annotations emphasizing geometric structure, color attributes, and motion characteristics.}
    \label{fig:data_pipeline}
\vspace{-3ex}
\end{figure*}

\section{Related Work}\label{sec:related}

\textbf{Vector Animation} offers a lightweight, resolution-independent representation that encodes appearance and temporal dynamics via structured shapes, effects, and control parameters. Early SVG-based methods relied on simple interpolation~\cite{DeepSVG,SVGFusion}, which limits the motion expressiveness. Later approaches incorporated external motion priors from reference GIFs~\cite{WW} or Text-to-video models~\cite{AniClipart,BLI,show1,videotetris}, and commercial tools such as LottieFiles Motion Copilot automate keyframe generation to ease manual workflows. However, these techniques decouple appearance and motion, relying on predefined dynamics or applying effects only to selected elements, preventing fully vectorized end-to-end animation synthesis.

We instead adopt the \textbf{Lottie} format, which unifies shapes, effects, and animations within a single JSON representation. Its simplicity allows modern VLMs to leverage strong instruction-following and multimodal reasoning, enabling \ours{} to learn end-to-end vector animation generation across diverse multimodal settings.

\noindent\textbf{Vector Graphics} are widely adopted in UI/UX, branding, and digital publishing. 
Prior work has proposed RNN-, VAE-, and transformer-based models~\cite{sketch-rnn,ClipGen,ClipVG,Im2Vec,DeepSVG,RSVG,StrokeNUWA,MES,SVGFusion,Iconshop,SuperSVG,SVGFormer} as well as diffusion-based approaches~\cite{SVGFusion,VGD,MARVEL} for SVG synthesis. With the rise of LLMs/VLMs~\cite{GPT-5,Gemini,Qwen,Qwen2.5VL}, autoregressive SVG generators~\cite{StrokeNUWA,LLM4SVG,OmniSVG} have demonstrated strong geometric and stylistic fidelity.  
Unlike static vector graphics, \textbf{vector animation} additionally models temporal and effect dynamics. \ours{} advances this line by jointly synthesizing appearance, effects, and motion in a unified autoregressive framework.

\noindent\textbf{Visual Autoregressive Generation.}\;
Autoregressive models treat visual content as token sequences and have been applied to images~\cite{llamagen,igpt,var,autoregressiveresi}, videos~\cite{ivideogpt, gr2,scalingvideo,lumos}, vector graphics~\cite{OmniSVG,starVector,LLM4SVG,svgbuilder}, and 3D~\cite{pointgpt,meshxl,motiongpt,shapegpt,g3pt,gr3}. Recent work replaces discrete heads with diffusion-based continuous predictors~\cite{mar,deng2024autoregressive}, and integrates pretrained VLMs for multimodal understanding~\cite{janus,emu3,ida,luenhancing,sdvlm}.  
We extend this paradigm to \textbf{vector animation}. By converting Lottie JSON into compact command–parameter sequences, we make complex animations amenable to sequence modeling. Built on a pretrained VLM~\cite{Qwen2.5VL}, \ours{} processes multimodal instructions and autoregressively generates Lottie commands, enabling controllable, high-fidelity vector animation synthesis.

\section{The MMLottie-2M Dataset}
\label{sec:MMLottie-2M dataset}

To support training and evaluation of multi-modal vector animation generation, we curate the first large-scale Lottie dataset, \textbf{MMLottie-2M}.
    In this section, we detail our data curation and processing pipeline, as well as a standardized evaluation protocol, \textbf{MMLottie-Bench} for the multi-modal generation tasks.

\begin{figure*}[h]
    \centering
    \includegraphics[width=\linewidth]{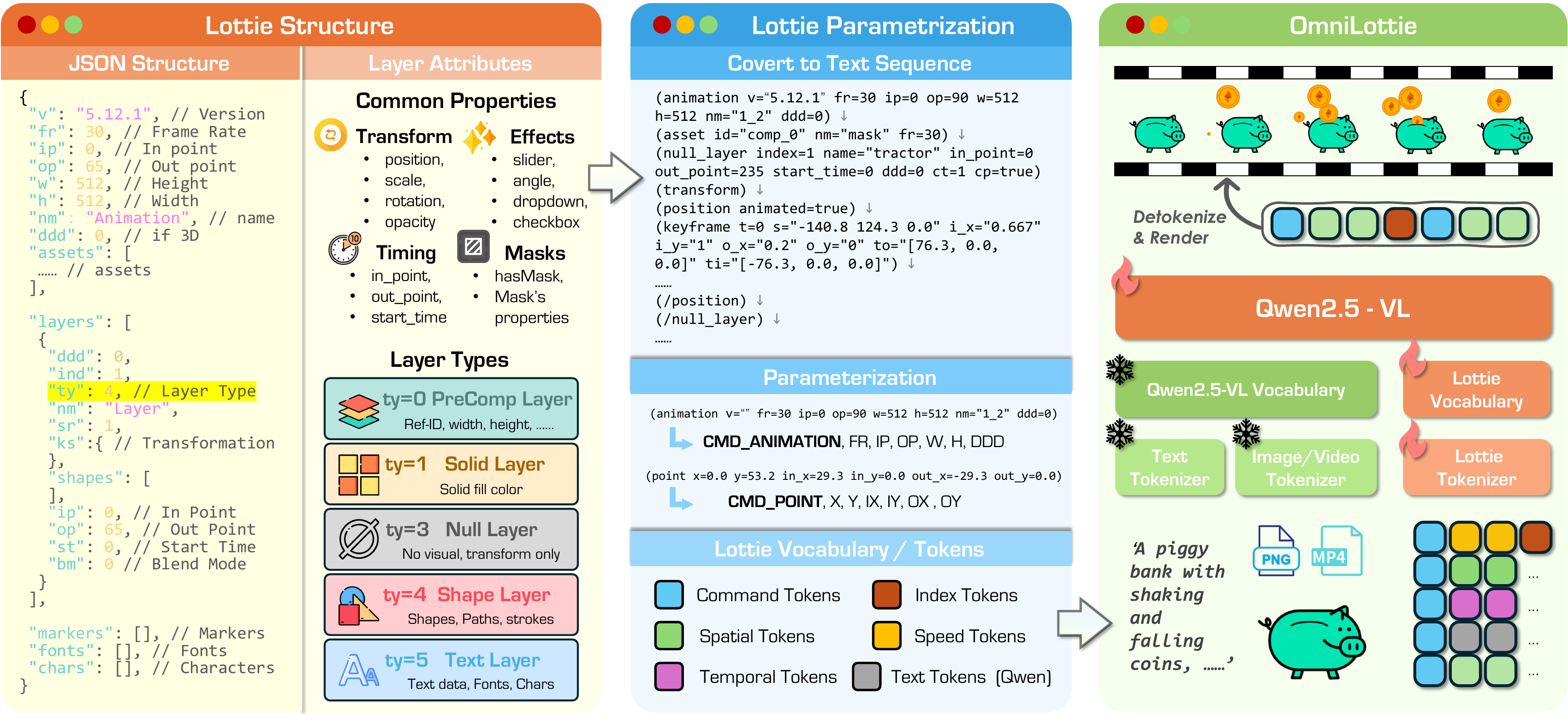}
    \vspace{-4ex}
    \caption{
\small \textbf{Overview of \ours}. \highlighttext{white}{myorange}{\textbf{Lottie Structure}}~: We reorganize the Lottie JSON representation, with a particular focus on the structure of its layers, including both common layer attributes and five special layer types. \highlighttext{white}{myblue}{\textbf{Lottie Parametrization}}~: The hierarchical JSON format of Lottie is flattened into a sequence of function calls, which are further parameterized to define a dedicated vocabulary and token set for Lottie. \highlighttext{white}{mygreen}{\textbf{\ours}}~: Built upon this parameterization, \ours~extends Qwen2.5-VL with a new tokenizer and vocabulary for Lottie, and is trained on our curated Lottie dataset.
}
\label{fig:pipeline}
\vspace{-3ex}
\end{figure*}

\subsection{MMLottie-2M}

\noindent \textbf{Lottie Data Curation.}
We construct the \textbf{MMLottie-2M} dataset by collecting Lottie animations from several major online platforms, including LottieFiles 
, IconScout 
, Flaticon 
, Iconfont 
, and Icons8. Web-crawled Lottie files often include irrelevant elements, such as base64 image layers, non-visual layers (audio, camera), and After Effects expressions that dynamically modify properties. These components complicate parameterization and are non-essential for rendering. We therefore remove such elements and discard files containing non-parameterizable layers to ensure clean, fully parameterizable data.

\noindent \textbf{SVG-Lottie Data Curation.} Direct learning from Lottie files is challenging due to their complex JSON structure, intricate layer indexing, and entangled visual and motion semantics. To simplify training, we generate an auxiliary Lottie dataset by animating static SVGs with predefined motions (e.g., translations, zooms, rotations), based on the large-scale OmniSVG~\cite{OmniSVG} collection. This decouples content from motion, improving alignment between visual components and animation conditions.

\noindent \textbf{Data Normalization.} To standardize training and evaluation procedures, all Lottie JSON files are normalized spatially to a $512\times512$ canvas and temporally to a 0–16 timestamp range, with center alignment preserving aspect ratio for non-square animations. 

\noindent \textbf{Motion Library for Data Augmentation.}    To mitigate the scarcity of native Lottie animations and close the distribution gap between static SVG conversions and authentic motion graphics, we develop a motion transfer pipeline. We analyze 1M native Lottie files to extract transform trajectories (rotation, scale, position, opacity) across keyframes, generating motion signatures that encode temporal patterns (e.g., ``fade-in + upward motion + scale-down''). By clustering semantically similar patterns with their caption keywords, we extract numerous canonical motion templates. These templates are then applied to static SVG-derived Lottie files through automated keyframe injection, yielding a massive scale of potential training instances. This approach drastically minimizes the path distribution gap while significantly broadening animated layer coverage, substantially improving the model's exposure to diverse motion dynamics without requiring additional manual animation efforts.

\noindent \textbf{Lottie Data Annotation.}

To support practical vector animation generation, we focus on three tasks: text-to-, image-text-to-, and video-to-Lottie. Lottie files are rendered into videos with random light-colored backgrounds for VLM-based annotation. Due to frame limitations, we adopt a coarse-to-fine strategy, the VLM first generates a brief overall caption describing subjects, objects, motion, color, and style, then provides finer temporal details across frames using cues like ``begins with'' and ``then.'' Keywords emphasizing geometry and motion are highlighted to improve text-following accuracy (\cref{fig:data_pipeline}).
For \textbf{Text-Image-to-Lottie}, a single frame is randomly selected from the rendered video, and the VLM is prompted to focus on foreground object motions. For \textbf{Video-to-Lottie}, the rendered videos directly serve as multi-modal instructions, simplifying annotation.

\subsection{MMLottie-Bench}

The lack of mature and standardized benchmarks and metrics for vector animation generation poses significant challenges in evaluating
(1) the quality of generated vector animations and
(2) the extent to which generators faithfully follow multi-modal instructions.
To address these issues, we introduce \textbf{MMLottie-Bench}, a comprehensive evaluation protocol for multi-modal vector animation generation.
Our goal is to construct a benchmark that
(1) reliably reflects a model’s practical utility in real-world scenarios and
(2) avoids the train--test overlap that commonly arises in conventional dataset splits.

To this end, we construct an evaluation set, termed the \textbf{Real Subset}, consisting of 450 samples curated from artist-designed Lottie animations collected from professional designers.
All evaluation samples are strictly disjoint from the training data, ensuring assessment on genuinely unseen, real-world content.
Specifically, for the \textit{Text-to-Lottie} task, we include 150 textual prompts derived from real Lottie animations.
For the \textit{Text-Image-to-Lottie} task, we extract one rendered frame together with its corresponding textual description from each of 150 real Lottie samples.
For the \textit{Video-to-Lottie} task, we include 150 rendered animation videos.

In addition, to further ensure the fairness and long-term robustness of our benchmark, particularly to mitigate potential contamination from future models trained on overly similar data, we construct a complementary \textbf{Synthetic Subset} via instruction-based synthesis using state-of-the-art generative models.
Concretely, for the \textit{Text-to-Lottie} task, we synthesize 150 textual prompts for each Lottie subset using GPT-4o.
For the \textit{Text-Image-to-Lottie} task, we synthesize 150 vector-style images using Gemini3.1-Pro Image, along with corresponding motion descriptions for evaluation.
For the \textit{Video-to-Lottie} task, we generate 150 additional shape descriptions paired with motion descriptions, which are then fed into Seedance~1.0 to produce reference videos.
To enhance the transparency of the benchmark, we record and release the entire data synthesis process for the Synthetic Subset.

MMLottie-Bench evaluates models along two key dimensions: \textbf{visual quality} and \textbf{semantic alignment} with the provided multi-modal inputs.

\section{Proposed Method: \ours}
\label{sec:OmniLottie}

\ours~consists of a well-designed Lottie tokenizer that encodes/decodes Lottie JSON files to/from compact sequences of discrete tokens, and a well-trained VLM~\cite{Qwen2.5VL} that can process images, videos, and texts as interleaved multi-modal instructions for auto-regressive Lottie generation.
The generated Lottie tokens are then detokenized to Lottie JSONs for vector animations as illustrated in \cref{fig:pipeline}.

\subsection{Preliminary: Lottie Layer Properties}

Lottie layer attributes can be grouped into three categories:

\noindent\textbf{Base layer properties} capture fundamental metadata, including identifiers (\textit{nm}, \textit{mn}), layer type \textit{ty}, indexing and hierarchy (\textit{ind}, \textit{parent}), and flags such as \textit{ddd} (3D) and \textit{hd} (hidden). Timing attributes (\textit{sr}, \textit{ip}, \textit{op}, \textit{st}) define each layer’s placement along the animation timeline.

\noindent\textbf{Visual layer properties} govern appearance and rendering behavior, including geometric transforms (\textit{ks}), orientation (\textit{ao}), matte relations (\textit{tt}, \textit{tp}, \textit{td}), masks (\textit{hasMask}, \textit{masksProperties}), effects (\textit{ef}), styles (\textit{sy}), blend mode (\textit{bm}), and rendering tags (\textit{cl}, \textit{ln}). The \textit{ct} flag controls transformation collapsing.

\noindent\textbf{Specific layer properties} depend on layer type: \textit{shapes} for shape layers; \textit{refId} for linking precomp, image, audio, and data assets; dimensions (\textit{w}, \textit{h}) for precomp clipping or solid-layer sizing; \textit{tm} for time remapping; \textit{t} for text content; and solid-layer parameters (\textit{sw}, \textit{sh}, \textit{sc}) defining width, height, and color.

\subsection{OmniLottie} 

\textbf{Lottie Structure.} Contemporary VLMs~\cite{GPT-5,Qwen2.5VL,Gemini} can generate JSON directly, but producing full Lottie JSON is inefficient for vector animation synthesis. The format contains extensive, largely invariant structural metadata, causing models to waste capacity on reproducing formatting tokens instead of learning animation-relevant priors. This redundancy increases sequence length and distracts the model from semantically important elements such as shapes, effects, and temporal dynamics.

To mitigate this, we reorganize the Lottie representation into a set of core properties (v, fr, ip, op, w, h, nm, ddd, layers) and conditional properties (assets, markers, fonts, chars), which appear only when required by specific layer types. Absent conditional fields are assigned empty values, yielding a more concise and structured Lottie representation.

\noindent \textbf{Lottie Tokenizer} abstracts Lottie animations into compact sequences of animation commands and control parameters, removing redundant metadata while preserving full generative flexibility. The tokenizer supports five fundamental layer types, each identified by a unique type parameter: Precomposition (ty=0), Solid (ty=1), Null (ty=3), Shape (ty=4), and Text (ty=5). Each layer type requires specialized parsing strategies to maintain structural integrity and temporal accuracy.

\begin{algorithm}[t]
\caption{\small Lottie Tokenizer}
\label{alg:tokenization}
\KwIn{Lottie JSON $\mathcal{J}$; Type-specific scales $\{s_t\}$, offsets $\{o_t\}$; Pretrained tokenizer $\mathcal{V}_{\text{text}}$}
\KwOut{Token sequence $\mathcal{T}$ (encoding) / Command sequence $\mathcal{S}$ (decoding)}

\SetKwProg{Fn}{Function}{:}{}

\tcp{Encoding}
\Fn{\textsc{Encode}($\mathcal{J}$)}{
    $\mathcal{M}, \{\mathcal{L}_i\} \gets \text{Parse}(\mathcal{J})$\;
    $\mathcal{T} \gets [\texttt{<META>}] + \text{Quantize}(\mathcal{M})$\;
    \ForEach{layer $\mathcal{L}_i$ with type $\tau_i$}{
        $\mathcal{T}.\text{append}([\texttt{<LAYER-}\tau_i\texttt{>}])$\;
        \ForEach{param $p$ in $\mathcal{L}_i$}{
            $\mathcal{T}.\text{append}(\lfloor p \cdot s_t \rfloor + o_t)$ \tcp*{Quantization}
        }
        \ForEach{text field $G$}{
            $\text{toks} \gets \mathcal{V}_{\text{text}}(G)$; \quad $\mathcal{T}.\text{extend}([\text{len}(\text{toks})] + \text{toks})$\;
        }
        $\mathcal{T}.\text{append}([\texttt{<END>}])$\;
    }
    \Return{$\mathcal{T}$}\;
}

\tcp{Decoding}
\Fn{\textsc{Decode}($\mathcal{T}$)}{
    $\mathcal{S} \gets []$\;
    \ForEach{command segment in $\mathcal{T}$}{
        $C \gets \text{new Command}()$\;
        \ForEach{numeric token $\text{tok}$}{
            $C.\text{add}((\text{tok} - o_t) / s_t)$ 
        }
        \ForEach{text token group}{
            $C.\text{add}(\mathcal{V}_{\text{text}}^{-1}(\text{tokens}))$\;
        }
        $\mathcal{S}.\text{append}(C)$\;
    }
    \Return{$\mathcal{S}$}\;
}
\end{algorithm}

Based on the distinct attributes of each layer type, Lottie tokenizer encapsulates the input JSON into fundamental properties (including v, ip, op, h, w, nm, ddd) and five categorized Python objects corresponding to different layer attribute classifications. We formalize Lottie animation as a structured hierarchy:

\begin{equation}
\mathcal{L} = \{\mathcal{M}, \mathcal{L}_1, \mathcal{L}_2, ..., \mathcal{L}_N\},
\end{equation}

\noindent where $\mathcal{M} = \{v, fr, ip, op, w, h, nm, ddd\}$ represents base metadata, and $\mathcal{L}_i$ denotes individual layers. 
Each layer is parameterized by its type $\tau \in \{0,1,3,4,5\}$ and attributes information $\mathcal{A}_{\tau}$:

\begin{equation}
\mathcal{L}_i = (\tau_i, \mathcal{A}_{\tau_i}, \mathcal{T}_i, \mathcal{E}_i),
\end{equation}
\noindent where $\mathcal{T}_i$ and $\mathcal{E}_i$ represent transformations and effects respectively. Thereby, Lottie tokenizer reconstructs the complete scene hierarchy through parent-child relationships, yielding a tree-structured representation that provides a lossless compression of the JSON format. This parameterized representation serves as the foundation for generating a compact, semantically rich token sequence.

The conversion systematically traverses each Lottie layer, serializing common attributes followed by type-specific properties into sequential tokens. By ignoring low-level JSON formatting, the model focuses on essential generative content. The Lottie Tokenizer formalizes this process (Algorithm~\ref{alg:tokenization}), parsing metadata first, then processing each layer with its type and attributes. As shown in \cref{fig:pipeline}, fundamental attributes are serialized as \texttt{animation v="..."}, and null layers are similarly encoded, yielding a structured, hierarchical text representation suitable for efficient VLM processing.

To adapt pretrained VLMs for Lottie generation, we map Lottie's hierarchical structure into a unified discrete vocabulary using an offset-based scheme. Distinct ranges are assigned to parameter types (temporal, spatial, transformations, and styles), avoiding token conflicts while preserving semantic coherence.

Our offset-based tokenization maps continuous Lottie parameters to discrete tokens:
\begin{equation}
\text{token}(x, t) = \lfloor x \cdot s_t \rfloor + o_t, 
\end{equation}

\noindent where $x$ is the parameter value, $t$ is the parameter type, $s_t$ is the type-specific scale factor, and $o_t$ is the vocabulary offset. The complete discrete token representation is:
\begin{equation}
\mathcal{T} = [\text{CMD}_1, p_{1,1}, ..., p_{1,k_1}, \text{CMD}_2, ..., \text{CMD}_M, p_{M,k_M}].
\end{equation}

\begin{table*}[t]
\vspace{-2ex}
\small
\caption{\small \textbf{Quantitative Evaluations.} We provide a comprehensive quantitative comparison between \textbf{Ours} and baseline methods across both \textit{Real subset} and \textit{Synthetic subset}. The bold numbers and underlined numbers represents the best and second best performance repectively. \textbf{Obj.} and \textbf{Motion} stand for object alignment and motion alignment.}
\label{tab:quantitative_sota}
\vspace{-2mm}
\centering
\setlength{\extrarowheight}{0pt}
\addtolength{\extrarowheight}{\aboverulesep}
\addtolength{\extrarowheight}{\belowrulesep}
\setlength{\aboverulesep}{0pt}
\setlength{\belowrulesep}{0pt}

\resizebox{0.98\linewidth}{!}{
\begin{tabular}{c|c|ccccccccccc}
\toprule
\textbf{Subset} & \textbf{Task} & \textbf{Methods} & \textbf{Time(s)} & \textbf{\# Tokens} & \textbf{Success Rate} & \textbf{FVD$\downarrow$} & \textbf{CLIP$\uparrow$} & \textbf{Obj.$\uparrow$} & \textbf{Motion$\uparrow$} & \textbf{PSNR$\uparrow$} & \textbf{SSIM$\uparrow$} & \textbf{DINO$\uparrow$} \\
\midrule

\rowcolor{blue!3} 
\cellcolor{white} & & DeepSeekV3 & 43.40 & 2.3k & 9.3\% & 671.80 & 0.2677 & 1.51 & 2.09 & - & - & - \\
\rowcolor{blue!3} 
\cellcolor{white} & & Qwen2.5-VL(3B) & 27.97 & 0.5k & 0.0\% & - & - & - & - & - & - & - \\
\rowcolor{blue!3} 
\cellcolor{white} & & GPT-5 & 43.40 & 1.4k & 12.7\% & 715.73 & 0.2600 & 0.73 & 0.71 & - & - & - \\
\rowcolor{blue!3} 
\cellcolor{white} & & Recraft & - & 54.1k & 77.3\% & \underline{300.70} & \textbf{0.2950} & \textbf{4.70} & \underline{4.68} & - & - & - \\
\rowcolor{blue!3} 
\cellcolor{white} & \multirow{-5}{*}[-0.2em]{\textbf{Text-to-Lottie}} & \textbf{Ours} & 33.71 & 21.2k & 88.3\% & \textbf{202.14} & \underline{0.2748} & \underline{4.44} & \textbf{5.94} & - & - & - \\
\cmidrule{2-13}
\rowcolor[RGB]{240,248,253} 
\cellcolor{white} & & Qwen2.5-VL(3B) & 33.60 & 0.4k & 0.0\% & - & - & - & - & - & - & - \\
\rowcolor[RGB]{240,248,253} 
\cellcolor{white} & & GPT-5 & 31.18 & 1.5k & 28.0\% & 546.65 & 0.2557 & 1.18 & 0.95 & - & - & - \\
\rowcolor[RGB]{240,248,253} 
\cellcolor{white} & & AniClipart & 1212.34 & - & 87.3\% & \underline{266.46} & \textbf{0.2935} & \underline{4.51} & \underline{3.47} & - & - & - \\
\rowcolor[RGB]{240,248,253} 
\cellcolor{white} & & Livesketch & 723.23 & - & 91.3\% & 868.18 & 0.2309 & 2.84 & 2.42 & - & - & - \\
\rowcolor[RGB]{240,248,253} 
\cellcolor{white} & \multirow{-5}{*}[-0.2em]{\textbf{Text-Image-to-Lottie}} & \textbf{Ours} & 88.57 & 23.4k & 93.3\% & \textbf{180.27} & \underline{0.2666} & \textbf{5.10} & \textbf{4.44} & - & - & - \\
\cmidrule{2-13}
\rowcolor[RGB]{253,248,247} 
\cellcolor{white} & & Qwen2.5-VL(3B) & 49.31 & 1.0k & 0.0\% & - & - & - & - & - & - & - \\
\rowcolor[RGB]{253,248,247} 
\cellcolor{white} & & GPT-5 & 45.61 & 1.1k & 9.2\% & \underline{639.13} & - & - & - & 13.34 & \underline{0.81} & 0.80 \\
\rowcolor[RGB]{253,248,247} 
\cellcolor{white} & & Gemini3.1-Pro & 16.19 & 1.0k & 0.0\% & 1076.22 & - & - & - & \underline{14.54} & 0.79 & \underline{0.88} \\
\rowcolor[RGB]{253,248,247} 
\cellcolor{white}\multirow{-14}{*}{\rotatebox{90}{\textbf{Real Subset}}} & \multirow{-4}{*}[-0.2em]{\textbf{Video-to-Lottie}} & \textbf{Ours} & 110.77 & 36.8k & 88.1\% & \textbf{227.11} & - & - & - & \textbf{16.08} & \textbf{0.82} & \textbf{0.92} \\

\midrule
\rowcolor{blue!3} 
\cellcolor{white} & & DeepSeekV3 & 56.71 & 2.3k & 7.4\% & 483.11 & 0.2677 & 1.43 & 1.98 & - & - & - \\
\rowcolor{blue!3} 
\cellcolor{white} & & Qwen2.5-VL(3B) & 94.36 & 0.4k & 0.0\% & - & - & - & - & - & - & - \\
\rowcolor{blue!3} 
\cellcolor{white} & & GPT-5 & 57.59 & 0.9k & 8.8\% & 637.29 & 0.2600 & 0.45 & 0.66 & - & - & - \\
\rowcolor{blue!3} 
\cellcolor{white} & & Recraft & - & 50.8k & 77.3\% & \underline{438.97} & \textbf{0.2950} & \underline{4.33} & \underline{3.12} & - & - & - \\
\rowcolor{blue!3} 
\cellcolor{white} & \multirow{-5}{*}[-0.2em]{\textbf{Text-to-Lottie}} & \textbf{Ours} & 37.93 & 13.4k & 82.1\% & \textbf{206.35} & \underline{0.2748} & \underline{4.31} & \textbf{5.63} & - & - & - \\
\cmidrule{2-13}
\rowcolor[RGB]{240,248,253} 
\cellcolor{white} & & Qwen2.5-VL(3B) & 31.06 & 0.3k & 0.0\% & - & - & - & - & - & - & - \\
\rowcolor[RGB]{240,248,253} 
\cellcolor{white} & & GPT-5 & 37.80 & 1.2k & 22.0\% & 560.11 & 0.2557 & 1.02 & 0.66 & - & - & - \\
\rowcolor[RGB]{240,248,253} 
\cellcolor{white} & & AniClipart & 1123.24 & - & 88.7\% & \underline{308.54} & \textbf{0.2935} & \underline{4.11} & \underline{2.79} & - & - & - \\
\rowcolor[RGB]{240,248,253} 
\cellcolor{white} & & Livesketch & 742.23 & - & 91.9\% & 1058.32 & 0.2309 & 2.01 & 1.91 & - & - & - \\
\rowcolor[RGB]{240,248,253} 
\cellcolor{white} & \multirow{-5}{*}[-0.2em]{\textbf{Text-Image-to-Lottie}} & \textbf{Ours} & 84.80 & 16.3k & 92.9\% & \textbf{225.45} & \underline{0.2666} & \textbf{4.44} & \textbf{3.98} & - & - & - \\
\cmidrule{2-13}
\rowcolor[RGB]{253,248,247} 
\cellcolor{white} & & Qwen2.5-VL(3B) & 42.19 & 1.1k & 0.0\% & - & - & - & - & - & - & - \\
\rowcolor[RGB]{253,248,247} 
\cellcolor{white} & & GPT-5 & 26.26 & 0.9k & 7.4\% & 576.52 & - & - & - & 13.33 & 0.71 & 0.78 \\
\rowcolor[RGB]{253,248,247} 
\cellcolor{white} & & Gemini3.1-Pro & 13.77 & 1.3k & 0.0\% & 1550.65 & - & - & - & \underline{13.89} & \underline{0.75} & \underline{0.83} \\
\rowcolor[RGB]{253,248,247} 
\cellcolor{white}\multirow{-14}{*}{\rotatebox{90}{\textbf{Synthetic Subset}}} & \multirow{-4}{*}[-0.2em]{\textbf{Video-to-Lottie}} & \textbf{Ours} & 109.53 & 41.4k & 80.7\% & 342.65 & - & - & - & \textbf{15.76} & \textbf{0.79} & \textbf{0.88} \\

\bottomrule
\end{tabular}
}
\vspace{-2mm}
\end{table*}

\noindent \textbf{Model architecture.} Building upon the token sequences $\mathcal{T}$ generated through the offset-based tokenization scheme, we train our model to learn the underlying distribution of Lottie animations represented in this discrete vocabulary space. To process interleaved multi-modal instructions, we adopt a pre-trained VLM, specifically Qwen2.5-VL~\cite{Qwen2.5VL}, as the backbone for \ours. We introduce an additional set of randomly initialized Lottie vocabulary embeddings to the model to represent the command and parameter structures produced by the Lottie tokenizer. Since the Qwen2.5-VL backbone is extensively pre-trained on diverse VLM tasks for concise response generation conditioned on the interleaved video, image, and text instructions, the pre-trained weights provide an effective initialization for generating Lotties adhering to multi-modal instructions.

\noindent \textbf{Training Objective.}  Similar to training LLMs, we train \ours~to generate new Lottie tokens $x_{s}^{[i]}$ conditioned on all previous tokens $x_{s}^{[<i]}$, as well as the provided multi-modal instructions $x_{c}$, with the standard cross-entropy loss.

\begin{equation}
\theta^{*} = \arg \min_{\theta} - \sum_{i = 1}^{L} \log P\left(x_{s}^{[i]} \vert x_{c}; x_{s}^{[<i]}; \theta\right)
\end{equation}

\section{Experiments}\label{sec:experiments}

In this section, we first introduce the baselines and evaluation metrics (\cref{subsec: baselines and metrics}).
We then present quantitative and qualitative evaluations in \cref{subsec: exp-comparison_quantitative} and \cref{subsec: exp-comparison_qualitative}, respectively. We conduct ablation studies in \cref{subsec: exp-ablation} to analyze the contribution of each component in \ours.

\begin{figure*}
    \vspace{-2ex}
    \centering
    \includegraphics[width=1\linewidth]{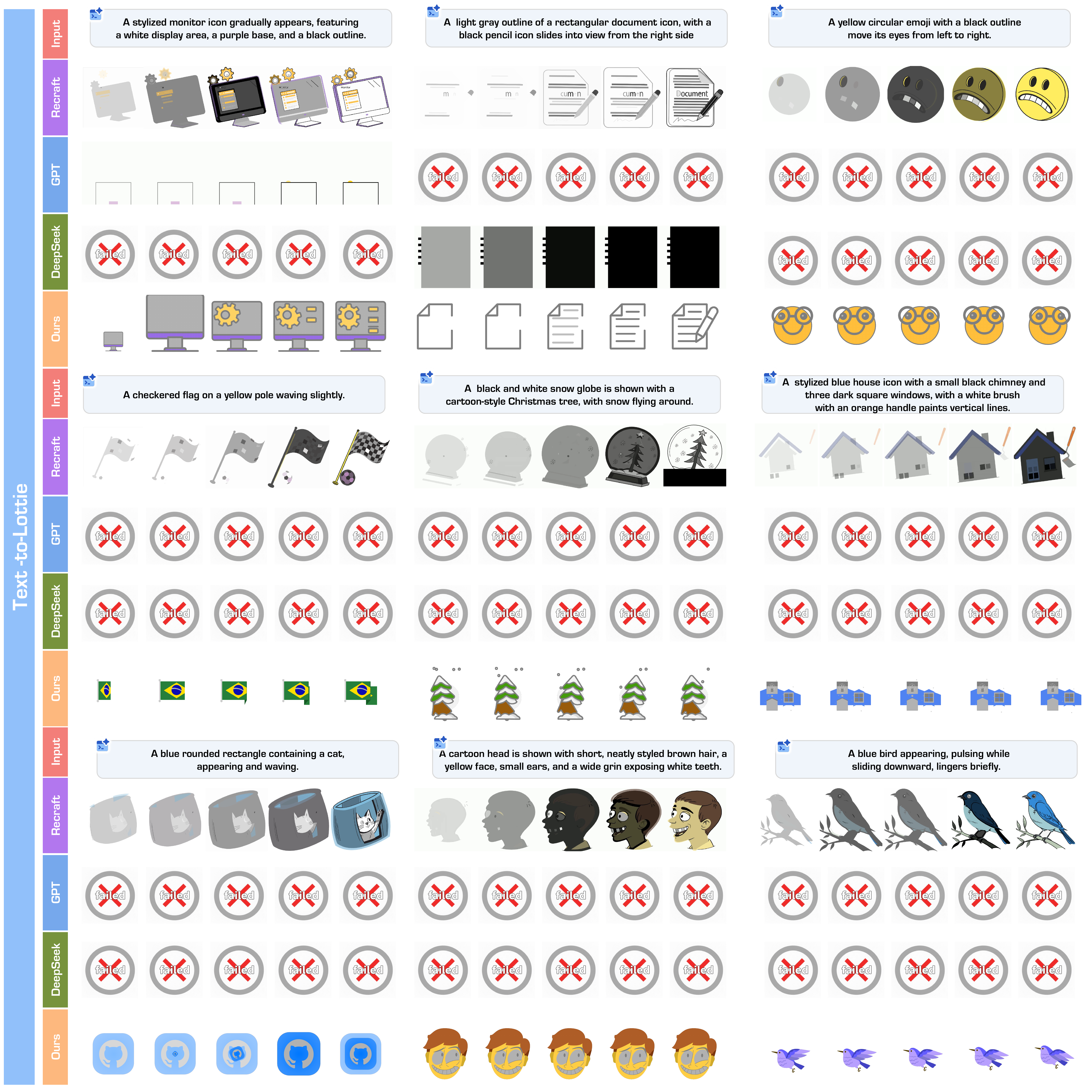}
\caption{\small \textbf{Qualitative comparison of Text-to-Lottie generation.}  Methods that failed to generate valid animations are omitted to ensure a clear comparison.}
\label{fig:qualitative_t2l}
\end{figure*}
\begin{figure*}
    \vspace{-2ex}
    \centering
    \includegraphics[width=1\linewidth]{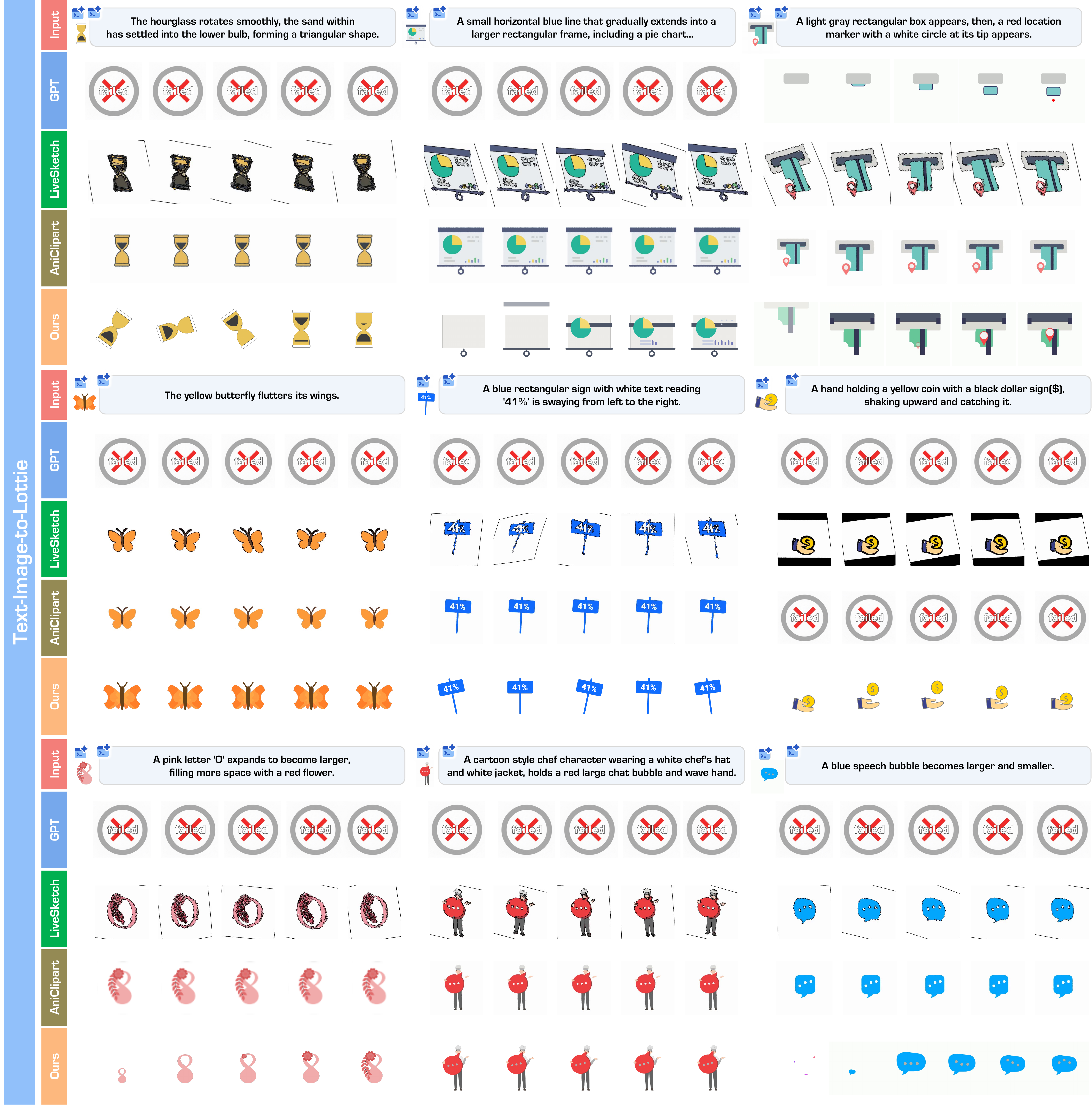}
\caption{\small \textbf{Qualitative comparison of Text-Image-to-Lottie generation.}  Methods that failed to generate valid animations are omitted to ensure a clear comparison.}
    \label{fig:qualitative_i2l}
\end{figure*}

\begin{figure*}
    \vspace{-2ex}
    \centering
    \includegraphics[width=1\linewidth]{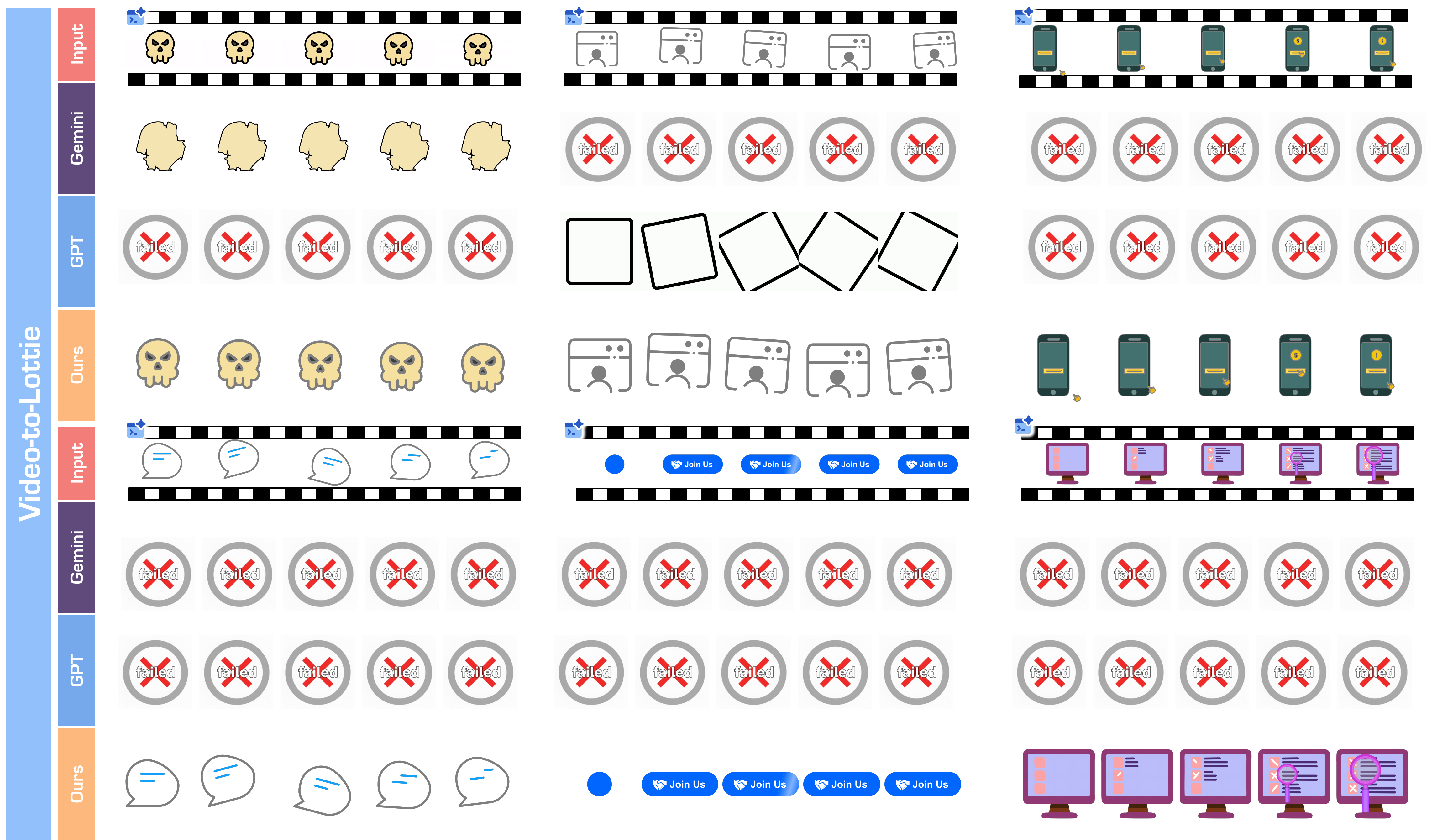}
\caption{\small \textbf{Qualitative comparison of Video-to-Lottie generation.}  Methods that failed to generate valid animations are omitted to ensure a clear comparison.}
    \label{fig:qualitative_v2l}
\end{figure*}

\subsection{Baselines and Evaluation Metrics}
\label{subsec: baselines and metrics}

\textbf{Baselines.}\;
For Text-to-Lottie, we compare \ours{} with DeepSeek~\cite{DeepSeekV3}, Qwen2.5-VL~\cite{Qwen2.5VL}, GPT-5~\cite{GPT-5}, and the commercial tool Recraft.  
For Text-Image-to-Lottie, we benchmark against Qwen2.5-VL, GPT-5, AniClipart~\cite{AniClipart}, and Livesketch~\cite{livesketch}.  
For Video-to-Lottie, we evaluate against Qwen2.5-VL, Gemini3.1-Pro~\cite{gemini3pro}, and GPT-5. All baselines use official implementations with recommended hyperparameters and preprocessing/postprocessing.

\noindent\textbf{Evaluation Metrics.}\;
For Text-to-Lottie and Text-Image-to-Lottie, we report FVD~\citep{FVD} for visual quality and CLIP similarity~\citep{CLIP} between prompts and rendered animation frames. We further introduce two caption-consistency metrics:  
\textbf{Object Alignment (Obj. Align)} (0–10), measuring object presence, type, count, visual traits, and spatial relations; and  
\textbf{Motion Alignment (Motion Align)} (0–10), assessing correctness of motion type, direction, magnitude, target objects, and smoothness, independent of object accuracy.  
Both metrics use Claude-3.5-Sonnet~\cite{claude} as an LLM judge; invalid generations are omitted, and blank outputs receive 0.

For Video-to-Lottie, we report FVD, PSNR~\cite{psnr}, SSIM~\cite{psnr}, and DINO~\cite{DINO}. Model efficiency is evaluated by (1) token efficiency average token length of generated Lottie JSON using the Qwen2.5-VL tokenizer and (2) computational cost average generation time per sample, with closed-source timings including the full API latency for realistic comparison.

\subsection{Quantitative Evaluation}
\label{subsec: exp-comparison_quantitative}

We report quantitative results for all methods in Tab.~\ref{tab:quantitative_sota} across Text-to-Lottie, Text-Image-to-Lottie, and Video-to-Lottie tasks. \ours~ delivers consistently superior performance on all metrics. In Text-to-Lottie generation, it achieves near-perfect success rates, the best FVD, and the strongest motion alignment. Although commercial tools like Recraft obtain competitive object alignment and CLIP scores, they lag behind in token efficiency and motion fidelity, while open-source language models show very low or zero success rates.

For Text-Image-to-Lottie, \ours{} again ranks first in FVD, object alignment, and motion alignment, while maintaining high reliability. Methods such as AniClipart and Livesketch, despite reasonable CLIP scores when successful, exhibit low success rates and significantly longer runtimes, limiting their practical utility.

In Video-to-Lottie, \ours{} preserves temporal and structural fidelity most effectively, achieving the best FVD, PSNR, SSIM, and DINO scores. GPT-5 provides moderate reconstruction quality but with notably lower success rates. Furthermore, \ours{} produces substantially richer token sequences than all baselines, enabling more expressive and detailed vector animation generation.

\subsection{Qualitative Evaluation} \label{subsec: exp-comparison_qualitative}

The qualitative comparisons in \cref{fig:qualitative_t2l,fig:qualitative_i2l,fig:qualitative_v2l} show that \ours{} delivers markedly better results across Text-to-Lottie, Text-Image-to-Lottie, and Video-to-Lottie tasks.  
For Text-to-Lottie, \ours{} achieves strong prompt alignment and high visual fidelity. In contrast, the commercial tool Recraft though capable of producing complex vector graphics—exhibits repetitive, poorly aligned motion, while GPT-5 suffers from high failure rates and low-quality Lottie outputs even when successful.

For Text-Image-to-Lottie, VLM-based approaches such as GPT-5 frequently fail and show weak alignment to both the reference image and text prompt. Optimization-based methods (AniClipart, LiveSketch) require long processing times, depend on raster-to-vector conversion before animation, cannot output native vector formats, and yield suboptimal visual results. \ours{} generates true vector animations with much shorter latency, high success rates, and superior visual quality.

For Video-to-Lottie, Qwen2.5-VL and Gemini fail to produce valid outputs, and GPT-5 yields occasional but low-quality conversions. In contrast, \ours{} produces high-quality vector animations that capture both motion patterns and visual characteristics from the input videos.

\begin{figure}
  \centering
    \centering
    \includegraphics[width=1\linewidth]{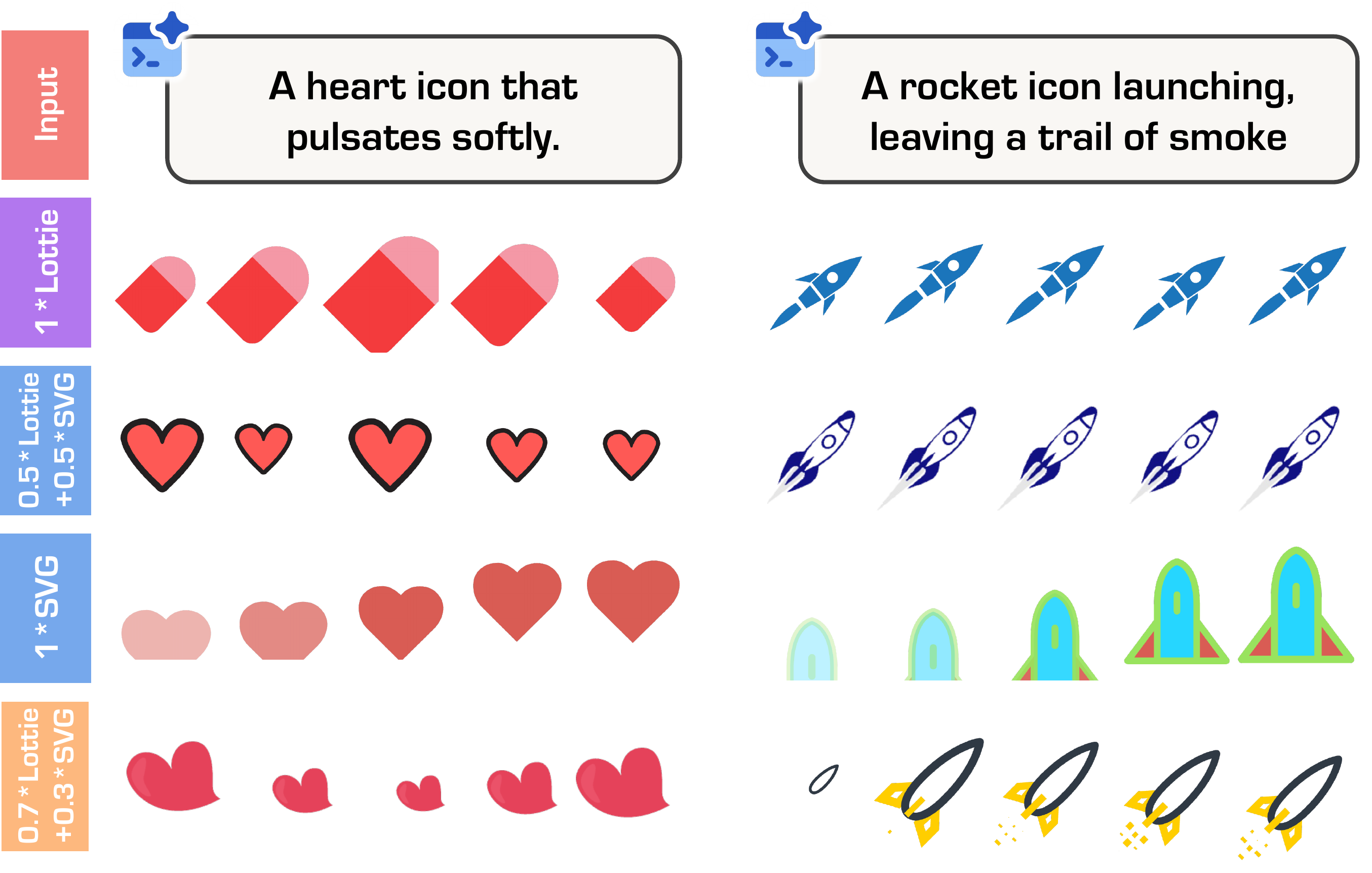}
    \caption{\small \textbf{Qualitative Ablation on Data Composition}. 
    }
    \label{fig:ablation of the dataset}
\end{figure}

\subsection{Ablation Studies}
\label{subsec: exp-ablation}

\noindent\textbf{Impact of SVG Data.}\; We evaluate different Lottie–SVG mixing ratios in ablation studies (\cref{tab:svg_ablation}, \cref{fig:ablation of the dataset}), including pure Lottie, Lottie with high SVG proportion, pure SVG, and Lottie with moderate SVG proportion (our final choice). Results show that moderate mixing yields the best performance on Text-to-Lottie and Text-Image-to-Lottie tasks. Lottie provides complex geometry and motion, while SVG contributes diverse shapes with simpler motion. Co-training improves geometric understanding, but excessive SVG biases the model toward simplistic motion, reducing Motion Alignment scores. Moderate mixing balances geometric richness with motion complexity, achieving optimal overall performance.

\begin{table}[t]
\centering
\caption{\small \textbf{Ablation Study on SVG Data Mixing.} We evaluate different combinations of Lottie~\lottie~and SVG~\svg~training data. The optimal performance is achieved with 30\% SVG data, which enhances geometric understanding while maintaining motion complexity. \textbf{Obj.} stands for object alignment.}
\label{tab:svg_ablation}
\resizebox{\linewidth}{!}{
\begin{tabular}{lcccccccc}
\toprule
\multirow{3}{*}[0.3em]{\textbf{Training Data}} & \multicolumn{4}{c}{\textbf{Text-to-Lottie}} & \multicolumn{4}{c}{\textbf{Text-Image-to-Lottie}} \\
\cmidrule(lr){2-5} \cmidrule(lr){6-9}
& \textbf{FVD}$\downarrow$ & \textbf{CLIP}$\uparrow$ & \textbf{Obj.} & \textbf{Motion} & \textbf{FVD}$\downarrow$ & \textbf{CLIP}$\uparrow$ & \textbf{Obj.} & \textbf{Motion} \\
\midrule
1*\lottie & 305.57 & 0.2573 & 1.85 & 4.92 & 405.39 & 0.2611 & 1.64 & 3.32 \\
0.5\lottie+0.5\svg & 285.22 & 0.2715 & 4.12 & 3.38 & 380.44 & 0.2642 & 3.91 & 3.25 \\
1*\svg & 342.61 & 0.2308 & 4.02 & 2.35 & 441.83 & 0.2552 & 3.88 & 2.63 \\
0.7*\lottie+0.3\svg & \textbf{269.50} & \textbf{0.2748} & \textbf{4.31} & \textbf{5.63} & \textbf{359.56} & \textbf{0.2666} & \textbf{4.10} & \textbf{3.44} \\
\bottomrule
\end{tabular}
}
\end{table}

\begin{figure}
  \centering
    \centering
    \includegraphics[width=0.9\linewidth]{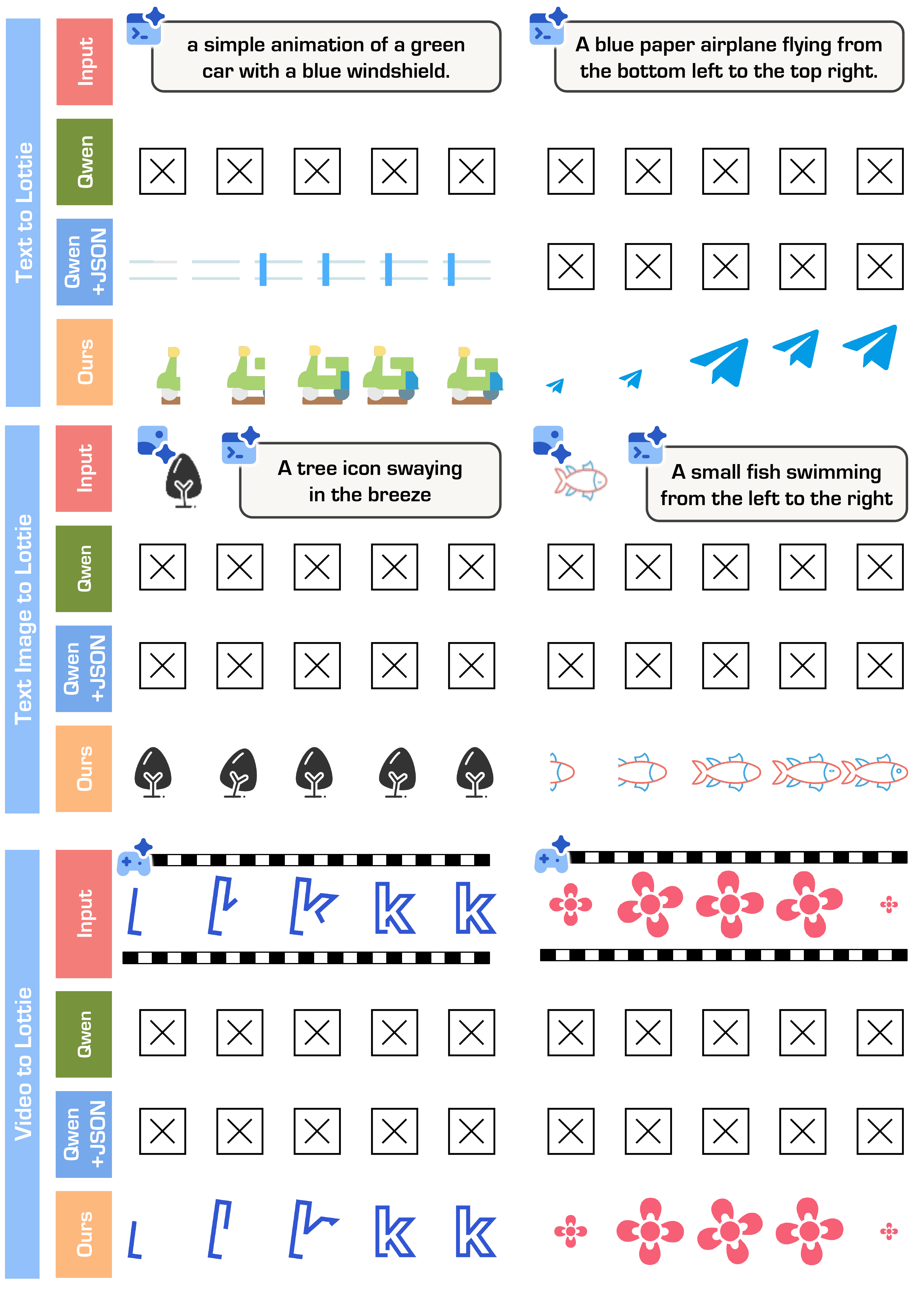}
    \caption{\small \textbf{Qualitative Ablation Study on Lottie Tokenizer}. }
    \label{fig:ablation of the tokenizer}
\end{figure}

\noindent\textbf{Necessity of the Lottie Tokenizer.}\;
We evaluate the impact of our Lottie Tokenizer via ablations on Text-to-, Text-Image-to-, and Video-to-Lottie tasks. Three configurations of Qwen2.5-VL are compared: pretrained, fine-tuned on raw JSON, and fine-tuned with our tokenizer. Metrics include FVD, CLIP, object and motion alignment for text/image tasks, and FVD, PSNR, SSIM, DINO for video. As shown in \cref{tab:abalation of the Lottie Tokenizer.} and \cref{fig:ablation of the tokenizer}, our tokenizer consistently improves performance across all tasks and metrics, demonstrating that its structured, compact representation enhances generation quality and computational efficiency.

\begin{table}[t]
\centering
\caption{\small \textbf{Ablation on Lottie Tokenizer}. Comparing pretrained Qwen2.5-VL~\qwen, fine-tuned on raw Lottie JSON~\lottiejson(including SVG-Lottie data), and trained with our tokenizer~\tokenizer. \textbf{SR} is the abbreviation of success rate, and \textbf{Obj.} stands for object alignment.}
\label{tab:abalation of the Lottie Tokenizer.}
\resizebox{\linewidth}{!}{
\begin{tabular}{cccccccccc}
\toprule
\textbf{Task} & \textbf{Setting} & \textbf{SR$\uparrow$} & \textbf{FVD$\downarrow$} & \textbf{CLIP$\uparrow$} & \textbf{Obj.$\uparrow$} & \textbf{Motion$\uparrow$} & \textbf{PSNR$\uparrow$} & \textbf{SSIM$\uparrow$} & \textbf{DINO$\uparrow$} \\
\midrule
& \qwen & 0.00\% & - & - & - & - & - & - & - \\
& \qwen\lottiejson & 13.4\% & 459.39 & 0.2600 & 1.33 & 1.82 & - & - & - \\
\multirow{-3}{*}[-2pt]{\textbf{T2Lottie}} & \qwen\lottiejson\tokenizer & 97.3\% & \textbf{269.50} & \textbf{0.2748} & \textbf{4.31} & \textbf{5.63} & - & - & - \\
\midrule
& \qwen &  0.00\% & - & - & - & - & - & - & - \\
& \qwen\lottiejson & 15.9\% & 476.23 & 0.2400 & 1.27 & 0.98 & - & - & - \\
\multirow{-3}{*}{\textbf{TI2Lottie}} & \qwen\lottiejson\tokenizer & 92.0\% & \textbf{359.56} & \textbf{0.2666} & \textbf{4.10} & \textbf{3.44} & - & - & - \\
\midrule
& \qwen &  0.00\% & - & - & - & - & - & - & - \\
& \qwen\lottiejson & 10.2\% & 431.23 & - & - & - & 17.23 & 0.76 & 0.68 \\
\multirow{-3}{*}{\textbf{V2Lottie}} & \qwen\lottiejson\tokenizer & 90.7\% & \textbf{281.95} & - & - & - & \textbf{19.05} & \textbf{0.91} & \textbf{0.88} \\
\bottomrule
\end{tabular}
}
\vspace{-2ex}
\end{table}


\section{Conclusions and Limitations}\label{sec:conclusions}
\noindent \textbf{Conclusion.} We introduce \textbf{\ours}, a unified framework for multi-modal vector animation generation. Leveraging our large-scale \textbf{MMLottie-2M} dataset, \ours~handles Text-, Text-Image-, and Video-to-Lottie tasks. Combined with the \textbf{Lottie tokenizer} and a pretrained VLM, it efficiently processes multi-modal inputs using compact token sequences. Experiments show that \ours~produces high-quality animations with strong visual fidelity and semantic alignment.

\noindent\textbf{Limitation.} While \ours~achieves higher success rates than direct JSON learning, autoregressive decoding can still yield invalid sequences. Generalization to diverse scenarios and context length remain challenges, limiting handling of complex animations. Future work could explore constrained decoding, reinforcement learning with renderability rewards, or agentic integration with professional tools like After Effects to improve reliability and practical applicability.

\newpage
\section{Disclaimer}

This \ourdataset (the ``Dataset'') is provided for research and non-commercial purposes only.
The Dataset is compiled from content that was originally publicly available on third-party websites. All copyrights, trademarks, and other intellectual property rights in the original content remain with their respective owners. The Dataset does not imply any endorsement, authorization, or license from the original rights holders.
The Dataset has been processed, filtered, and reorganized by the authors. Such processing does not alter the ownership or intellectual property status of the underlying content.
The Dataset is provided “as is” and “as available,” without warranties of any kind, express or implied, including but not limited to warranties of accuracy, completeness, reliability, merchantability, fitness for a particular purpose, or non-infringement.
The authors and the affiliated organization make no representations or guarantees regarding the legality, correctness, or appropriateness of the Dataset or its use in any jurisdiction. Users are solely responsible for ensuring that their use of the Dataset complies with all applicable laws, regulations, and third-party rights.
Under no circumstances shall the authors or the affiliated organization be liable for any claims, damages, losses, or other liabilities arising from or related to the use of the Dataset.

\newpage

\noindent\textbf{\LARGE Appendix}
\appendix

\maketitle

\setcounter{section}{0}
\makeatletter

\setcounter{table}{3} %
\setcounter{figure}{6} %

\vspace{8pt}

\noindent In this appendix, we provide comprehensive additional content organized as follows:

\vspace{4pt}

\begin{itemize}
\item \textbf{Section A}: Additional qualitative results across all three generation tasks, along with a systematic failure mode analysis covering our method and all baselines, including a unified five-level failure taxonomy and comparative cost-benefit analysis.

\item \textbf{Section B}: Complete user study details, including study design and data selection, participant recruitment, evaluation interface and procedure, and correlation analysis between automated metrics and human judgments.

\item \textbf{Section C}: In-depth technical details of the Lottie Tokenizer, including layer-specific parsing, vocabulary offset design with statistical analysis, sequence-to-token conversion algorithms, token-to-sequence reconstruction algorithms, design principles, and analysis of vector representation preservation and token efficiency.

\item \textbf{Section D}: Comprehensive MMLottie-2M dataset documentation, covering the complete five-stage data collection and processing pipeline, dataset statistics (source distribution, temporal and spatial statistics, structural complexity), caption statistics and semantic analysis, and evaluation protocol details including LLM-as-judge prompts.

\item \textbf{Section E}: Detailed Lottie layer type specifications, including layer type classification and comprehensive property documentation for all supported layer types.
\end{itemize}

\section{Qualitative Results and Failure Analysis}
\label{app:qualitative}

This appendix provides comprehensive qualitative evaluation through additional generation examples and systematic failure mode analysis.

\subsection{Additional Qualitative Results}
\label{app:additional_qualitative}

\begin{figure*}
    \centering
    \includegraphics[width=0.98\linewidth]{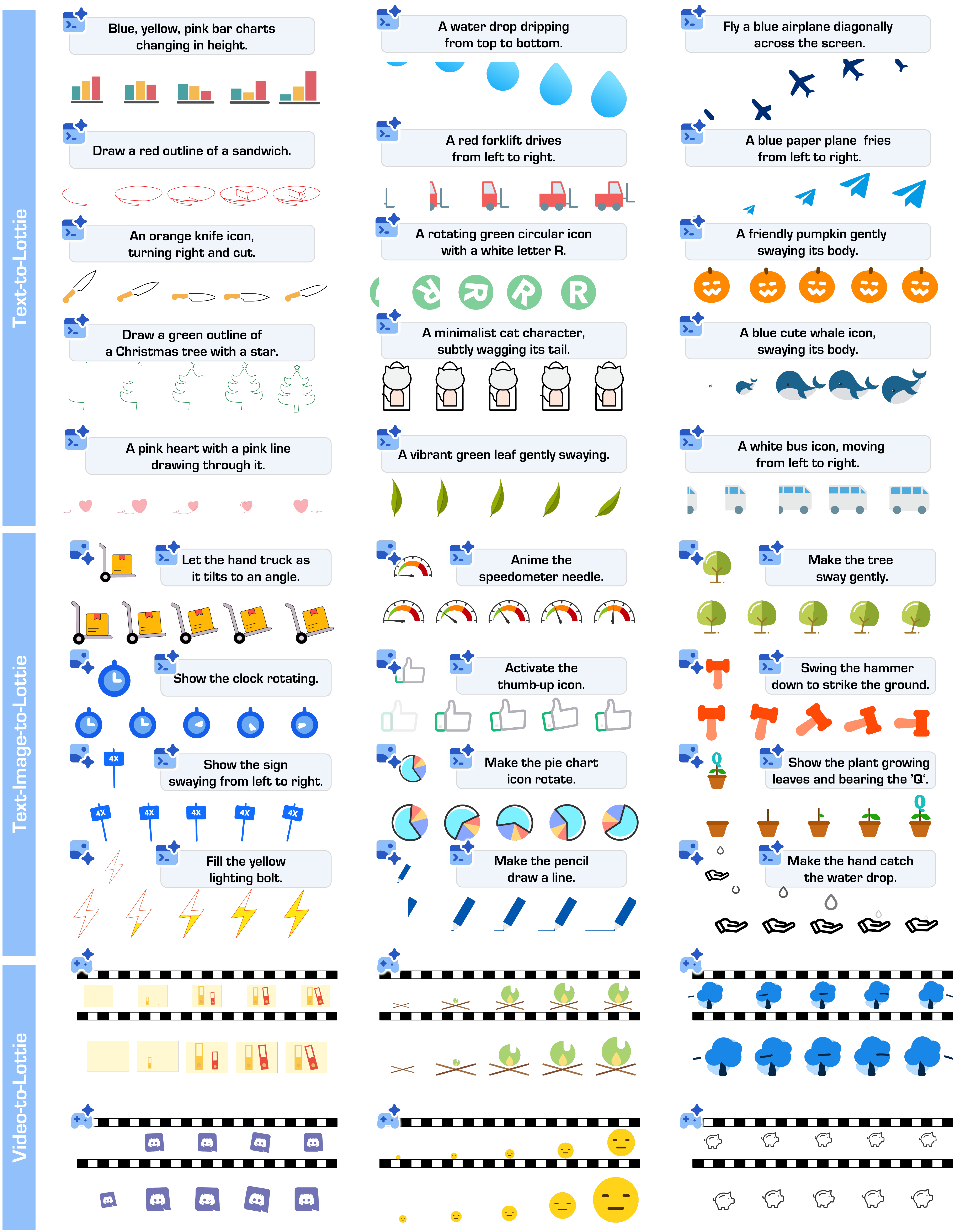}
    \caption{\small \textbf{Additional Qualitative Results Across Three Generation Tasks}. We present additional examples for Text-to-Lottie, Text-Image-to-Lottie, and Video-to-Lottie generation.}
    \label{fig:supple_vis}
\end{figure*}

We provide additional qualitative examples demonstrating the performance of \ours~across all three generation tasks: text-to-Lottie, text-image-to-Lottie, and video-to-Lottie. These examples showcase diverse scenarios including varying complexity levels, different object types, and various motion patterns. Results are shown in~\cref{fig:supple_vis}.

\subsection{Failure Case Analysis}
\label{app:failure_analysis}

Despite achieving high success rates across all tasks (97.3\%, 92.0\%, and 90.7\% for Text-to-Lottie, Text-Image-to-Lottie, and Video-to-Lottie respectively, understanding the failure modes of both \ours~and baseline methods is crucial for future improvements. We present a systematic analysis organized by method type, followed by a comparative summary.

\subsubsection{Failure Taxonomy Overview}

We categorize all failure modes into a unified hierarchy based on the stage at which failures occur: (1) \textbf{Level 1 - Specification Failures} where generated JSON violates Lottie schema requirements; (2) \textbf{Level 2 - Structural Failures} with valid schema but empty or malformed content; (3) \textbf{Level 3 - Rendering Failures} with valid structure but invisible or incorrect visual output; (4) \textbf{Level 4 - Pipeline Failures} with valid Lottie but downstream conversion errors; and (5) \textbf{Level 5 - Input-Dependency Failures} where method is inapplicable to given input characteristics.

\subsubsection{Failure Analysis of \ours}

\ours~primarily exhibits Level 2 and Level 3 failures, indicating strong specification compliance but occasional content generation issues. Failure rates increase with input complexity: Text-to-Lottie (2.7\%) $<$ Text-Image-to-Lottie (8.0\%) $<$ Video-to-Lottie (9.3\%).

\noindent\textbf{Structural Generation Failure ($\sim$35\% of failures).}
The model generates syntactically valid Lottie JSON with correct global structure (\texttt{v}, \texttt{fr}, \texttt{ip}, \texttt{op}, \texttt{w}, \texttt{h}), but produces an empty \texttt{layers} array. This typically occurs when text prompts are ambiguous or input images/videos contain complex visual elements that the model struggles to decompose into vector primitives.

\begin{lstlisting}
{
  "v": "5.12.1", 
  "fr": 25.0, 
  "ip": 0.0, 
  "op": 105.0,
  "w": 512.0, 
  "h": 512.0,
  "layers": [],  // No layer generated
  "assets": [], 
  "markers": []
}
\end{lstlisting}

\noindent\textbf{Rendering-Level Failures ($\sim$65\% of failures).}
Layers are generated but result in blank rendering outputs. We identify five primary sub-categories. First, missing style attributes ($\sim$40\%) occur when Lottie separates geometry and styling into sibling nodes, and the model generates path data (\texttt{ty:~"sh"}) but omits corresponding fill (\texttt{ty:~"fl"}) or stroke (\texttt{ty:~"st"}) nodes. Second, temporal visibility errors ($\sim$25\%) involve layers with \texttt{ip}/\texttt{op} values outside the animation duration (e.g., \texttt{ip:~100} when global \texttt{op:~60}) or imperceptibly short durations. Third, opacity/scale value errors ($\sim$25\%) arise because Lottie uses 0-100 for opacity and percentage-based scaling, causing predictions like \texttt{o:~1} (1\% opacity) or \texttt{s:~[1,1]} (1\% scale) to render objects effectively invisible. Fourth, off-canvas positioning ($\sim$10\%) places coordinates far outside canvas bounds (e.g., \texttt{p:~[5000,5000]} for a 512×512 canvas). Video-to-Lottie shows higher susceptibility to temporal errors due to motion extraction challenges, while Text-Image-to-Lottie exhibits more style-related failures from matching both textual semantics and visual appearance.

\begin{table}[h]
\small
\caption{\textbf{Failure Case Statistics.} Distribution of failure types across different tasks.}
\label{tab:failure_statistics}
\centering
\begin{tabular}{lccc}
\toprule
\textbf{Failure Type} & \textbf{T2L} & \textbf{TI2L} & \textbf{V2L} \\
\midrule
Category I: Empty Layers & 40.0\% & 30.0\% & 35.0\% \\
\midrule
\multicolumn{4}{l}{\textit{Category II: Invalid Rendering}} \\
\quad Missing Style & 30.0\% & 35.0\% & 20.0\% \\
\quad Temporal Errors & 10.0\% & 15.0\% & 25.0\% \\
\quad Opacity Issues & 10.0\% & 10.0\% & 10.0\% \\
\quad Scale Collapse & 5.0\% & 5.0\% & 5.0\% \\
\quad Off-Canvas & 5.0\% & 5.0\% & 5.0\% \\
\bottomrule
\end{tabular}
\vspace{-3mm}
\end{table}

\subsubsection{Failure Analysis of LLM/VLM Baselines}

LLM/VLM baselines demonstrate fundamentally different failure modes, primarily at Level 1 (specification) and Level 3 (rendering). Qwen2.5-VL~\cite{Qwen2.5VL} achieves 0.0\% success across all tasks due to schema hallucination, generating plausible-looking but entirely incompatible JSON that conflates Lottie with generic animation formats, using attributes like \texttt{"version": "2.0"} instead of \texttt{"v"}, \texttt{"frames"} instead of keyframe interpolation, and \texttt{"duration"} instead of \texttt{"op"}. DeepSeek~\cite{DeepSeekV3} achieves 29.3\% success but suffers from attribute hallucination, generating structurally valid JSON while injecting attributes into incorrect nodes, likely from SVG training data where geometry and styling are co-located. For example, it places color attributes \texttt{"c"} directly in rectangle shape nodes (\texttt{"ty": "rc"}) rather than in separate fill nodes (\texttt{"ty":"fl"}). Additionally, DeepSeek frequently generates text layers without corresponding \texttt{fonts} definitions.

GPT-5~\cite{GPT-5} (12.7\%-68.67\% success) and Gemini~\cite{Gemini} (0.0\% Video-to-Lottie) produce schema-valid JSON with diverse rendering errors including geometric degeneration where Bézier control points collapse to identical coordinates, transform stack errors from incorrect nesting causing coordinate system corruption, and keyframe discontinuities with non-monotonic time values or infinite tangents. Recraft achieves 100\% Lottie generation success but 22.7\% failure in Lottie-to-video rendering, revealing limitations of decoupled systems through unsupported features (expression-driven animations or blend modes incompatible with video renderers), performance bottlenecks (excessively complex layer hierarchies with $>$50 nested groups causing timeouts), and format mismatch (features valid in web players but incompatible with video export libraries).

\subsubsection{Failure Analysis of Optimization-Based Baselines}

AniClipart~\cite{AniClipart} and LiveSketch~\cite{livesketch} employ multi-stage optimization pipelines with text-to-video diffusion priors. Their failures (92.7\% and 52\% failure rates) occur primarily at Level 5 (input-dependency) and through optimization non-convergence.

\noindent\textbf{Input Feature Dependency Failures.}
AniClipart requires semantic keypoints (skeletal joints, object corners) to parameterize motion as Bézier curves, causing $\sim$85\% of its failures when abstract shapes or geometric patterns lack discernible keypoints, simple shapes yield insufficient keypoints for ARAP deformation constraints, or overlapping elements create keypoint ambiguity. This explains the 7.3\% success rate—the method is inherently inapplicable to most icon/clipart designs. LiveSketch, designed for sketch-style inputs with sparse strokes, experiences $\sim$60\% of its failures on photo-realistic images or gradient-filled shapes that cannot decompose into strokes, non-white backgrounds introducing artifacts during stroke extraction, and dense inputs causing the local-global motion optimization to overfit to noise.

\noindent\textbf{Optimization Non-Convergence ($\sim$30\% of combined failures).}
Both methods rely on Score Distillation Sampling (SDS) with highly non-convex optimization landscapes, leading to local minima convergence with degenerate solutions exhibiting minimal or jittering motion, mode collapse to generic motions (uniform scaling, simple translation) that ignore semantic structure, and gradient pathologies from differentiable rasterization. The pipeline architecture compounds errors across stages: PNG $\rightarrow$ SVG $\rightarrow$ Keypoint Extraction $\rightarrow$ SDS Optimization $\rightarrow$ Animated SVG $\rightarrow$ Lottie JSON, where vectorization errors, optimization timeouts (1200s for AniClipart, 780s for LiveSketch), and format conversion losses accumulate throughout.

\subsubsection{Comparative Summary}

The failure mode distribution reveals a clear hierarchy correlating with method design: Qwen2.5-VL at Level 1 (specification) achieves 0\% success; DeepSeek at Level 1 (attribute-level) achieves 29.3\% success; AniClipart at Level 5 (input-dependency) achieves 7.3\% success; LiveSketch at Level 5 (input-dependency) achieves 48.0\% success; GPT-5/Gemini at Level 3 (rendering) achieve 12.7-68.67\% success; Recraft at Level 4 (pipeline) achieves 77.3\% success; and \ours~at Level 2-3 (structure/rendering) achieves 90.7-97.3\% success.

The cost-benefit analysis further differentiates the methods. AniClipart requires 1200s with 7.3\% success, yielding 16,438s per successful animation. LiveSketch requires 780s with 48.0\% success, yielding 1,625s per successful animation. \ours~requires only 28.57s with 92.0\% success, yielding 31s per successful animation, a 52× speedup over LiveSketch and 530× over AniClipart per successful generation.

This analysis reveals that \ours's focused training on Lottie-specific data successfully addresses specification and structural challenges that plague general-purpose LLM/VLMs, while its direct generation paradigm avoids the input-dependency and optimization convergence issues inherent to optimization-based approaches. The remaining rendering-level failures in \ours~represent tractable improvements through enhanced training data coverage and numerical precision.

\section{User Study}
\label{app:user_study}

We conduct a comprehensive user study follows the procotal in prior arts~\cite{xu2025withanyone,chang2025oneig,zhuang2025vistorybench} to evaluate the quality and multi-modal alignment of generated Lottie animations. Twenty participants were recruited and asked to rank generated animations from different methods according to four criteria: visual quality, condition adherence, animation quality, and geometric fidelity. Each participant completed evaluation tasks across three generation scenarios: Text-to-Lottie, Text-Image-to-Lottie, and Video-to-Lottie. 
The results demonstrate that OmniLottie consistently achieves the highest average ranking across all dimensions and tasks, indicating superior generation quality and stronger alignment with multi-modal inputs. Moreover, correlation analysis reveals that our proposed metrics (Object Alignment and Motion Alignment) exhibit strong positive correlation with human judgments, validating their effectiveness as automated evaluation measures. Detailed study design, ranking protocols, and statistical analysis are provided in the following sections.

\subsection{Study Design and Data Selection}

Our user study evaluates model performance across three distinct generation tasks. Due to the practical constraints of participant workload and the necessity of ensuring high-quality evaluation samples, we adopt a stratified sampling strategy that balances task diversity with evaluation depth.

\begin{table*}[t]
\centering
\caption{\small \textbf{Participant Instructions and Evaluation Procedure for Lottie Animation Generation.}}
\label{tab:user_study_instructions}
\begin{tcolorbox}[
    title=Participant Instructions and Evaluation Procedure,
    enhanced,
    colback={rgb,255:red,249;green,250;blue,255},
    colframe={rgb,255:red,109;green,153;blue,255},
    fonttitle=\bfseries,
]

\textbf{Study Overview and Task Description}

This study evaluates different methods for generating vector animations (Lottie format) from multi-modal instructions. Three generation tasks are assessed: (1) Text-to-Lottie: generating animations from text descriptions, (2) Text-Image-to-Lottie: generating animations from both text and reference images, and (3) Video-to-Lottie: generating animations that reconstruct input videos. For each trial, you will observe 3-4 generated animations from different methods and rank them individually on a 5-star scale (5 = best, 1 = worst) across four evaluation dimensions. All generated animations are rendered as looping videos for consistent viewing. Please carefully review the input conditions before evaluating each animation.

\vspace{6pt}

\textbf{Evaluation Dimensions (Per-Animation Ranking)}

Rate each generated animation independently on the following four criteria:

\vspace{3pt}

\textbf{1. Visual Quality}

Assess the overall aesthetic quality and visual appeal of the animation. Consider rendering smoothness, color harmony, clarity of shapes, absence of visual artifacts, and professional appearance. Higher visual quality indicates better-rendered, more polished animations with clean vector graphics and pleasant visual composition.

\vspace{3pt}

\textbf{2. Condition Adherence}

Evaluate how well the generated animation matches the given input conditions. For Text-to-Lottie, assess whether the animation accurately reflects all elements described in the text prompt. For Text-Image-to-Lottie, evaluate alignment with both the reference image's visual content and the text description. For Video-to-Lottie, assess how faithfully the animation reconstructs the input video's content, structure, and appearance. Stronger adherence means better alignment between input conditions and generated output.

\vspace{3pt}

\textbf{3. Animation Quality}

Judge the quality of motion and temporal behavior. Consider smoothness of transitions, naturalness of movements, absence of jittering or abrupt changes, appropriate timing and pacing, and coherent temporal progression. Higher animation quality indicates smoother, more fluid, and more natural-looking motion patterns that are visually pleasing and temporally consistent.

\vspace{3pt}

\textbf{4. Geometric Fidelity}

Evaluate the accuracy and completeness of geometric shapes and structural elements. Consider shape accuracy, preservation of fine details, structural integrity, appropriate proportions, and absence of geometric distortions. For Text-Image-to-Lottie and Video-to-Lottie, also assess how well geometric elements from the input are preserved. Higher geometric fidelity indicates more accurate, detailed, and structurally sound vector representations.

\vspace{6pt}

\textbf{Important Evaluation Guidelines}

\begin{itemize}[leftmargin=*, itemsep=2pt, parsep=0pt]
\item \textbf{Independent Rating}: Evaluate each dimension separately. An animation may score high on one dimension but low on another.
\item \textbf{Absolute Quality}: Focus on the absolute quality of each animation rather than making relative comparisons between methods during rating.
\item \textbf{Full Scale Usage}: Use the complete 1-5 star range. Reserve 5 stars for excellent quality, 4 for good, 3 for acceptable, 2 for poor, and 1 for very poor quality.
\item \textbf{Consistency}: Apply the same standards across all evaluation groups to ensure consistent judgment.
\end{itemize}

\end{tcolorbox}
\end{table*}

\begin{table*}[t]
\centering
\caption{\small \textbf{Correlation Statistics Between Automated Metrics and Human Rankings.} Reported values include Pearson's $r$, Spearman's $\rho$, and Kendall's $\tau$ with corresponding $p$-values. Higher absolute correlation values indicate stronger agreement between automated metrics and human judgments.}
\label{tab:correlation}
\resizebox{0.75\linewidth}{!}{
\begin{tabular}{lccc}
\toprule
\textbf{Metric vs. Human Rating} & \textbf{Pearson $r$ ($p$)} & \textbf{Spearman $\rho$ ($p$)} & \textbf{Kendall $\tau$ ($p$)} \\
\midrule
Object Align vs. Geometric Fidelity & 0.4521 (3.21e-42) & 0.4638 (8.45e-45) & 0.3512 (2.18e-41) \\
Motion Align vs. Animation Quality & 0.4823 (1.52e-48) & 0.4917 (4.23e-51) & 0.3724 (5.67e-47) \\
CLIP vs. Condition Adherence & 0.3867 (6.78e-34) & 0.3945 (2.14e-35) & 0.2918 (1.43e-32) \\
FVD vs. Visual Quality & -0.3245 (4.89e-28) & -0.3312 (1.67e-29) & -0.2456 (3.21e-26) \\
\bottomrule
\end{tabular}
}
\end{table*}

\noindent \textbf{Sample Selection Strategy.} We conduct the user study on all 300 benchmark items for each of the three tasks in the MMLottie Benchmark (Text-to-Lottie, Text-Image-to-Lottie, and Video-to-Lottie), resulting in 300 generation cases per task. In total, 900 evaluation cases are included in the study (300 Text-to-Lottie, 300 Text-Image-to-Lottie, and 300 Video-to-Lottie). For cases where the generation fails to produce valid animations, the result is displayed as ``Failed'' to the evaluators, allowing participants to assess both generation quality and system reliability within a unified evaluation framework.

\noindent \textbf{Baseline Methods.} For each task, we compare OmniLottie against multiple baseline methods. Text-to-Lottie comparisons include Deepseek, Recraft, Qwen2.5-VL and GPT-5. Text-Image-to-Lottie comparisons include GPT-5, Aniclipart and Livesketch. Video-to-Lottie comparisons include GPT-5 and Gemini3.1-Pro. Each evaluation group presents 3-4 anonymized animations rendered from generated Lottie files, shown in randomized order to prevent bias.

\noindent \textbf{Participant Recruitment and Training.} We recruit 20 participants with basic understanding of vector graphics and animation concepts. Before the formal evaluation, all participants undergo a standardized training session using a tutorial set of 3 examples (one per task) with detailed explanations of evaluation criteria. This training ensures consistent understanding of the ranking dimensions and reduces inter-rater variability.

\subsection{Evaluation Interface and Procedure}

The evaluation is conducted through an online questionnaire platform where participants view the rendered animations and provide rankings. For each evaluation group, participants are shown: (1) the multi-modal input condition (text prompt, image+text, or video), (2) 3-4 generated animations rendered as looping videos with transparent or white backgrounds, and (3) detailed evaluation instructions. Participants then rank each generated animation individually on a 5-star scale (5 = best, 1 = worst) across four dimensions: visual quality, condition adherence, animation quality, and geometric fidelity. The ranking is performed independently for each dimension, allowing fine-grained assessment of different quality aspects.  We provide the participant instructions and evaluation procedure for Lottie animation generation in \cref{tab:user_study_instructions}.

\subsection{Correlation Analysis}

We analyze the correlation between automated metrics and human rankings to validate the effectiveness of our proposed evaluation measures. As shown in ~\cref{tab:correlation}, our Object Alignment and Motion Alignment metrics demonstrate strong positive correlation with corresponding human judgments across all three statistical measures (Pearson's $r$, Spearman's $\rho$, and Kendall's $\tau$), with highly significant $p$-values ($p < 0.001$). The Motion Alignment metric shows particularly strong correlation (Pearson $r = 0.4823$), indicating that it effectively captures perceptually meaningful animation quality. The CLIP score also correlates moderately with human-rated condition adherence, though with slightly lower correlation coefficients compared to our specialized metrics. These results validate that our proposed metrics provide reliable automated assessment aligned with human perception.

\section{Lottie Tokenizer: Technical Details}
\label{app:tokenizer_details}

\subsection{Vocabulary Offset Design}
\label{app:offset_scale}

To efficiently represent diverse Lottie animation parameters in a unified token space, we employ a systematic offset strategy that maps each parameter category to a distinct vocabulary region. This design prevents overlap between semantically different parameters while preserving their numerical relationships within categories.

\begin{figure*}[t]
    \centering
    \includegraphics[width=\linewidth]{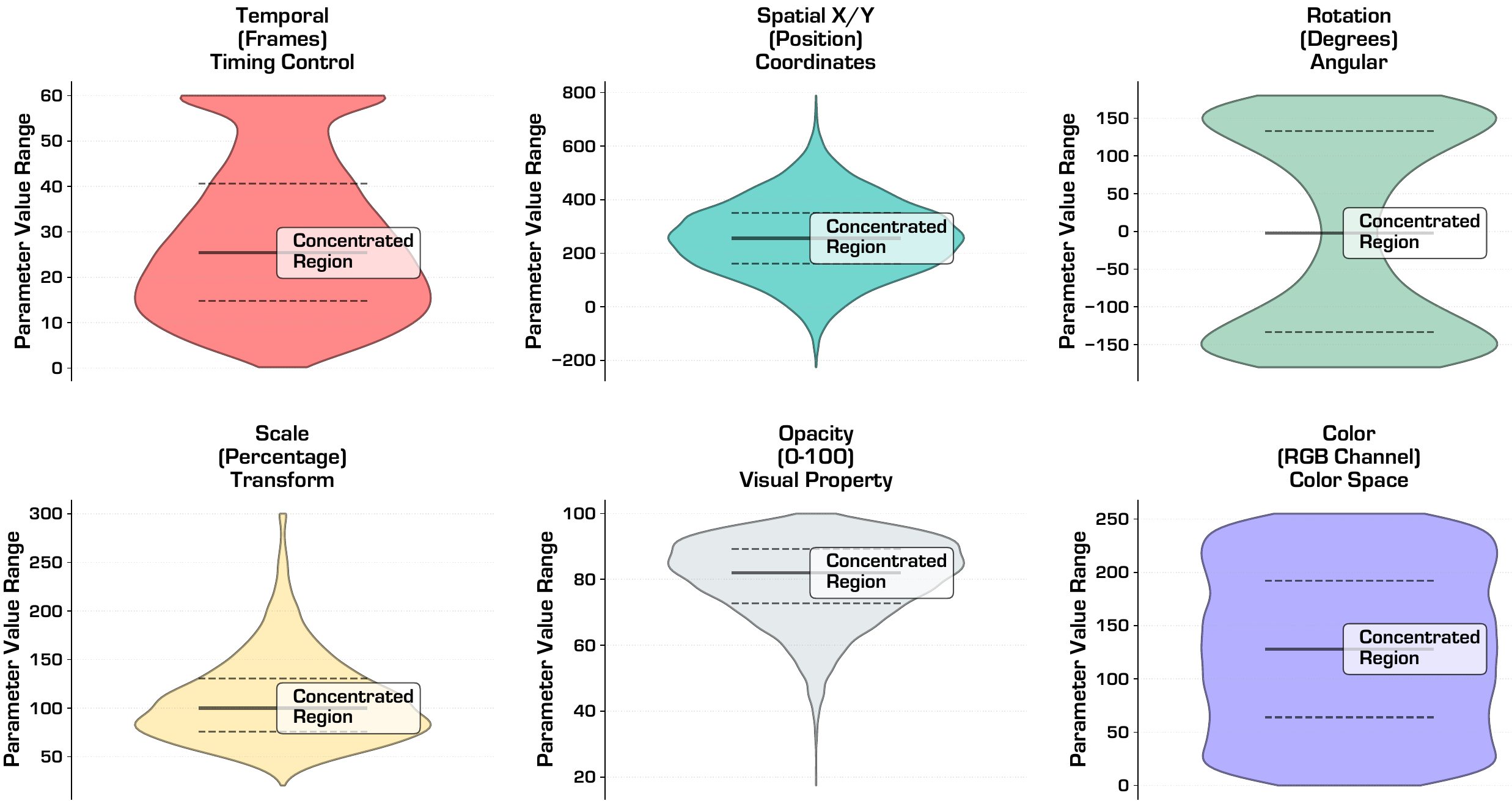}
    \caption{
        \textbf{Parameter Distribution Analysis for Vocabulary Offset Design.} 
        We analyze the distribution of six key parameter types across our entire dataset to determine optimal vocabulary offset ranges. 
        Each violin plot shows the probability density of parameter values, with solid horizontal lines indicating median values and dashed lines marking quartiles. 
        The shaded regions highlight concentrated value ranges where the majority of parameters cluster, guiding our offset allocation strategy.
    }
    \label{fig:offset_ranges}
\end{figure*}

The value ranges for each parameter type are determined through comprehensive statistical analysis of our entire dataset. We examine the distribution of parameter values across all animations and identify concentrated ranges by filtering extreme outliers that represent noise or edge cases. For example, spatial parameters such as positions and anchor points exhibit natural clustering within negative to positive coordinate ranges, as illustrated in ~\cref{fig:offset_ranges}. Similarly, rotation angles concentrate within full rotation cycles, while scale factors cluster around percentage-based transformations. This data-driven approach ensures our vocabulary efficiently covers the practical parameter space without allocating excessive tokens to rarely-used extreme values.

Our vocabulary design encompasses multiple parameter categories with varying complexity. Binary flags such as animation state indicators and boolean properties occupy minimal space with two possible values each. Small categorical enumerations including blend modes, line caps, and shape types require modest discrete ranges. Continuous numerical parameters span larger spaces, with temporal values controlling timing, spatial coordinates managing positioning, and transformation parameters governing scaling and rotation. Color channels map to standard intensity ranges, while specialized categories like easing curves and opacity values occupy intermediate scales. Tokenizer-based parameters for text and identifiers bypass offset transformation entirely, maintaining their original token representations. This hierarchical organization allocates vocabulary space proportionally to parameter complexity, with simple flags using minimal tokens and complex transformations requiring extensive ranges to capture subtle variations in animation behavior.

\subsection{Sequence to Token Conversion}
\label{app:seq2token}

After parsing Lottie JSON into command sequences, we discretize continuous parameters into a unified token vocabulary through our offset-based mechanism. This process handles three distinct parameter categories: continuous numerical values, text-based content, and structural markers.

\noindent\textbf{Numerical Parameter Discretization.} For continuous parameters (temporal, spatial, transformation, and style attributes), we apply type-specific scaling and offset transformations:
\begin{equation}
\text{token}(p, t) = \lfloor p \cdot s_t \rfloor + o_t
\end{equation}
where $p$ denotes the parameter value, $t$ represents its semantic type, $s_t$ is the scaling factor, and $o_t$ is the vocabulary offset. Each parameter type occupies a distinct vocabulary range determined by data distribution analysis, ensuring no token conflicts across categories.

\noindent\textbf{Text Content Tokenization.} Text-related attributes including font families, character strings, and textual identifiers require special handling to preserve the pretrained model's linguistic knowledge. Rather than converting text to numerical tokens, we employ the backbone VLM's native tokenizer (Qwen2.5-VL) to encode these elements. For commands involving text parameters (font, char, reference\_id), we adopt a structured representation:
\begin{equation}
\mathcal{T}_{\text{text}} = [\text{count}, \text{tok}_1, \text{tok}_2, ..., \text{tok}_n]
\end{equation}
where count indicates the number of tokens, followed by the actual token sequence. This design maintains the semantic richness of text embeddings while integrating seamlessly with numerical parameter tokens.

\noindent\textbf{Padding and Structural Markers.} To handle optional parameters and maintain structural information, we introduce special marker tokens. For each vocabulary range defined by offset $o_t$, we reserve a marker token below the valid range ($\text{start}(o_t) - 1$) to represent padding values. Command boundaries are explicitly marked with command-specific start and end tokens, enabling clear hierarchical structure in the token sequence:
\begin{equation}
\mathcal{S} = [\text{CMD}_i, p_{i,1}, ..., p_{i,k}, \text{END}_i, \text{CMD}_{i+1}, ...]
\end{equation}

The complete tokenization procedure is formalized in Algorithm~\ref{alg:tokenization}, which systematically processes each command and its parameters according to their semantic types.

\setcounter{algocf}{1}

\subsection{Token-to-Sequence Reconstruction}
\label{app:token2seq}

The detokenization process reconstructs valid Lottie JSON from generated token sequences through systematic parsing and reverse transformation. This involves three key operations: command boundary detection, parameter value recovery, and text content decoding.

\noindent\textbf{Command Structure Parsing.} The reconstruction begins by identifying command boundaries using start and end markers. Each command block is processed sequentially, with the command token determining the expected parameter structure. For tokenizer commands (containing text content), the parser first reads regular numeric parameters, then processes token groups by extracting the count value followed by the specified number of token IDs.

\noindent\textbf{Parameter Value Recovery.} Numerical parameters undergo inverse transformation to recover their original continuous values:
\begin{equation}
p = \frac{\text{token} - o_t}{s_t}
\end{equation}
Special marker tokens are recognized and converted back to PAD\_VAL placeholders, maintaining structural information during reconstruction. The parameter type $t$ is inferred from the command structure and parameter position, ensuring correct offset application.

\noindent\textbf{Text Content Decoding.} For text-based parameters, the token IDs are decoded using the same pretrained tokenizer employed during encoding:
\begin{equation}
\text{text} = \mathcal{V}_{\text{text}}^{-1}([\text{tok}_1, ..., \text{tok}_n])
\end{equation}
This preserves semantic consistency between encoding and decoding phases, maintaining the linguistic properties learned by the backbone VLM.

The complete reconstruction procedure is outlined in Algorithm~\ref{alg:tokenization}, which systematically processes the token sequence while maintaining hierarchical structure and parameter relationships. This deterministic mapping ensures that valid token sequences always reconstruct to renderable Lottie JSON files.

\subsection{Tokenization Design Principles}

Our tokenization framework incorporates several key design principles that collectively enable efficient and accurate vector animation generation:

\noindent\textbf{Separation of Concerns.} By treating command tokens, numerical parameters, and text content as distinct categories with specialized handling, we allow the model to focus its capacity on learning meaningful animation patterns rather than low-level syntax. Command tokens establish structural hierarchy, numerical parameters encode geometric and temporal attributes, and text tokens preserve semantic richness.

\noindent\textbf{Vocabulary Efficiency.} The offset-based mapping eliminates redundancy by assigning each parameter type to a non-overlapping vocabulary range proportional to its complexity. Binary flags occupy minimal space, while continuous transformations span broader ranges to capture subtle variations. This allocation strategy maximizes token utilization while maintaining sufficient precision for animation fidelity.

\noindent\textbf{Reconstruction Guarantees.} The deterministic mapping between parameters and tokens ensures lossless round-trip conversion. Special marker tokens explicitly represent padding and structural boundaries, enabling the decoder to unambiguously reconstruct the hierarchical Lottie structure. This property is critical for maintaining animation integrity throughout the generation process.

\noindent\textbf{Model Compatibility.} Integration with pretrained text tokenizers preserves the linguistic capabilities of backbone VLMs while extending them to the animation domain. Text parameters retain their original embeddings, allowing the model to leverage learned language understanding when processing font families, character content, and textual identifiers embedded within animation structures.

\subsection{Vector Representation Preservation and Token Efficiency}
\label{app:vector_preservation}

\begin{figure}[h]
\centering
\includegraphics[width=1\linewidth]{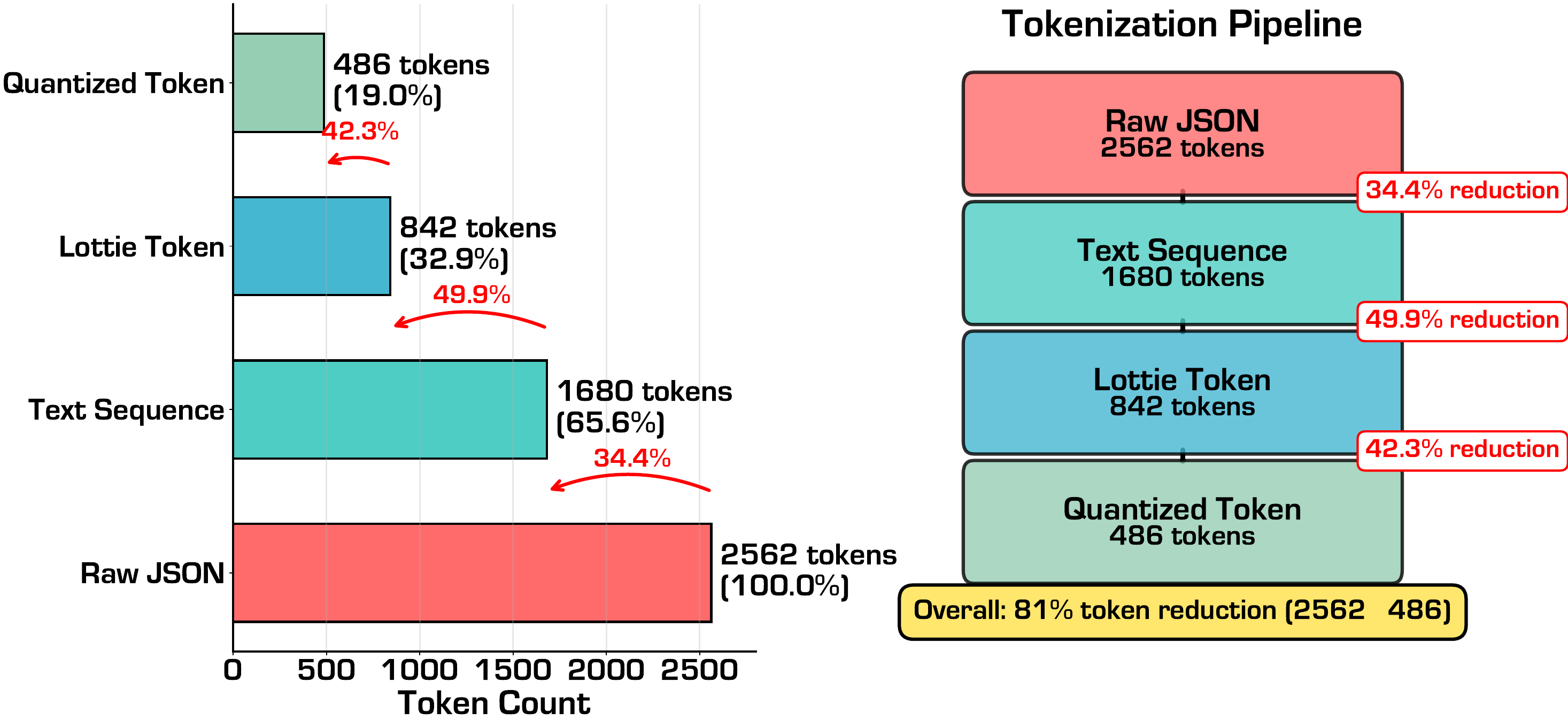}
\caption{\textbf{Token Efficiency Through Progressive Abstraction.} Our Lottie tokenizer progressively reduces sequence length from raw JSON (2,562 tokens) through structured text sequences (884 tokens) to our final command-based representation (486 tokens), achieving 5.3× compression. The discretization operates purely at the representation level and does not compromise the vector nature of the output, as continuous parameter values are faithfully recovered during detokenization.}
\label{fig:token_efficiency}
\end{figure}

A fundamental concern in our tokenization approach is whether the discrete token representation contradicts Lottie's inherent vector nature. We emphasize that our tokenization scheme does not compromise the vector characteristics of Lottie animations but rather provides a more efficient intermediate representation for learning and generation. The Lottie format itself, while being a pure vector representation, suffers from substantial redundancy when encoded as raw JSON due to verbose syntax, repeated structural markers, and metadata overhead. Our tokenization pipeline systematically addresses this inefficiency through three progressive transformations: first, converting JSON to structured text sequences by removing syntactic redundancy while preserving semantic content; second, organizing these sequences into command-parameter pairs that explicitly capture the functional structure of animation primitives; and third, applying offset-based discretization that maps continuous parameters into a compact vocabulary space. Critically, this entire process is fully reversible through deterministic detokenization, ensuring that every generated token sequence can be reconstructed into a valid, renderable Lottie JSON file that maintains complete vector fidelity, resolution independence, and editability. The token efficiency gains, as illustrated in \cref{fig:token_efficiency}, demonstrate that our approach reduces sequence length by 81\% compared to raw JSON (from 2,562 to 486 tokens on average) while enabling the model to focus its capacity on learning s animation patterns rather than formatting conventions. This design philosophy aligns with recent advances in structured representation learning, where appropriate abstraction layers can dramatically improve both training efficiency and generation quality without sacrificing the fundamental properties of the target domain.

\section{MMLottie-2M Dataset: Comprehensive Details}
\label{app:dataset_details}

\subsection{Data Collection and Processing Pipeline}

Our MMLottie-2M dataset construction involves two primary data sources: web-crawled Lottie animations and synthesized Lottie animations from static SVG files. Both data sources undergo a unified processing pipeline to ensure consistency and quality. The complete data processing pipeline  consists of five stages: data collection, Lottie cleaning, spatial-temporal normalization, video rendering, and multi-modal annotation.

\noindent\textbf{Stage 1: Data Collection.} We collect Lottie animations from two distinct sources. For web-crawled data, we aggregate animations from five major online platforms: LottieFiles, IconScout, Flaticon, Iconfont, and Icons8. These platforms host professionally designed animations covering diverse categories including user interface elements, icons, illustrations, and motion graphics. In total, we collect approximately 1.2 million Lottie files from these sources. For SVG-derived data, we leverage the large-scale OmniSVG~\cite{OmniSVG} collection, which contains over 2 million static SVG files. We convert these SVG files into static Lottie representations and augment them with procedurally generated animations. Specifically, we implement seven types of basic animations: horizontal translation (moving left/right), vertical translation (moving up/down), scaling (zoom in/out), rotation (clockwise/counterclockwise), opacity change (fade in/out), and combined motions. For each SVG file, we randomly select 1-3 animation types and apply them with randomized parameters (duration, direction, magnitude) to create diverse animated variations. This approach yields approximately 800,000 additional Lottie animations with decoupled visual content and motion semantics.

\noindent\textbf{Stage 2: Lottie Cleaning.} Raw Lottie files from web sources often contain elements that complicate parameterization or are non-essential for rendering. We systematically remove three types of problematic components. First, we eliminate base64-encoded image layers, which embed raster images directly in the JSON structure and introduce dependencies on external bitmap resources. Second, we remove non-visual layers including audio layers and camera layers, as these do not contribute to the 2D vector animation output. Third, we strip JavaScript/ECMAScript expressions embedded within visual layers, which follow the After Effects expression syntax and can dynamically modify property values at runtime. While these expressions provide convenience during the design process, they introduce non-deterministic behavior and complicate the learning of animation priors. After cleaning, we discard any Lottie files that still contain non-parameterizable layers such as 3D layers or data layers, ensuring that all retained animations are fully representable through our tokenization scheme. 

\noindent\textbf{Stage 3: Spatial-Temporal Normalization.} To standardize the training and evaluation workflow, we apply unified spatial and temporal normalization to all Lottie files from both sources. Spatially, we normalize each animation to a $512\times512$ pixel canvas. For Lottie files with non-square original resolutions, we apply center alignment to preserve aspect ratio and prevent geometric distortion. Specifically, if the original width $w$ and height $h$ differ, we scale the content to fit within the 512×512 canvas while maintaining the aspect ratio $r = \min(512/w, 512/h)$, then center the scaled content. This ensures that the entire original content remains visible without cropping. Temporally, we normalize all timestamp values to a unified range from 0 to 60. The original in-point ($ip$), out-point ($op$), and all keyframe timestamps are linearly scaled according to the formula: $t_{\text{norm}} = 60 \times (t_{\text{orig}} - ip) / (op - ip)$. This temporal normalization enables consistent sequence lengths across animations and facilitates batch training.

\noindent\textbf{Stage 4: Video Rendering.} To support multi-modal annotation and evaluation, we render each normalized Lottie file into a video sequence. We use the official Lottie renderer to generate MP4 videos at 30 frames per second with a resolution of $512\times512$ pixels. To help the model distinguish foreground vector content from backgrounds, we render each animation with a randomly selected solid light-colored background. The background colors are sampled from a palette of 20 pastel colors including light gray (\#F5F5F5), light blue (\#E3F2FD), light pink (\#FCE4EC), light yellow (\#FFF9C4), and light green (\#E8F5E9). For each Lottie file, we render the complete animation duration. Additionally, for the Text-Image-to-Lottie task, we extract a single keyframe from each rendered video. The keyframe is selected at a random timestamp $t \in [0.2, 0.8] \times T$, where $T$ is the total duration, to ensure we capture a representative moment rather than the initial or final static states. This process yields three types of rendered outputs for each Lottie file: (1) a complete video sequence, (2) a randomly selected keyframe image, and (3) the same keyframe image with a randomly added solid background.

\noindent\textbf{Stage 5: Multi-Modal Annotation.} We employ Qwen2.5-VL~\cite{Qwen2.5VL} to generate textual descriptions for each rendered video. Due to the frame number limitation of VLMs for video understanding (typically 8-16 frames), it is impractical for the model to provide comprehensive descriptions of all elements and motions in a single pass. Therefore, we adopt a coarse-to-fine annotation strategy where the VLM progressively describes video details at multiple levels of granularity. The annotation process follows a two-stage prompting strategy as detailed in \cref{tab:instruction_templates}. In the first stage, we prompt the VLM to generate a brief, one-sentence caption that roughly describes the overall content of the video, including coarse-level information about the main subject, objects, characters, motions, colors, and animation style. The prompt emphasizes that every visible element must be described with its color, and the model should use specific terms for visual effects such as ``fading in'', ``sliding out'', ``gaussian blur'', and ``drop shadow''. In the second stage, we instruct the model to examine different frames of the video and provide more detailed temporal descriptions using connective phrases such as ``begins with'', ``then'', ``the last'', and ``finish with''. This stage focuses on capturing motion progression, spatial relationships between elements, and the evolution of visual properties over time. To enhance the model's text-following capability, we particularly emphasize keywords related to geometric elements (e.g., ``circle'', ``rectangle'', ``star'') and motion descriptions (e.g., ``rotating clockwise'', ``scaling up'', ``translating left''). The final annotation for each Lottie file consists of both the concise overview caption and the detailed temporal description, providing flexibility for different generation tasks. For the Video-to-Lottie task, the rendered videos directly serve as multi-modal instructions without additional textual annotation.


\begin{table*}[hbtp]
\caption{\small \textbf{Instructions for Different Tasks}. Instructions including annotation, text-to-SVG, image-to-SVG and character-reference SVG generation.}

\centering

\begin{tcolorbox}[title=Instructions for Different Tasks,
enhanced,
skin first=enhanced,
skin middle=enhanced,
skin last=enhanced,
colback={rgb,255:red, 255; green, 250; blue, 250},
colframe={rgb,255:red, 227; green, 108; blue, 194},
fonttitle=\bfseries]

\begin{itemize}
\item \textbf{Employed Qwen2.5-VL for Video Captioning:} Please analyze this video and provide detailed factual descriptions. Follow these strict requirements:

IMPORTANT GUIDELINES:
\begin{enumerate}
\item Describe ALL visible elements with their COLORS: people, animals, objects, text, shapes, backgrounds, scenes, etc.
\item EVERY element must include color description (e.g., ``red text'', ``blue rectangle'', ``white background'', ``black outline'')
\item Include accurate counts when distinguishable (e.g., ``2 yellow cookies, 1 blue cup'')
\item Describe spatial relationships precisely (e.g., ``two red balloons above a gray iron table'')
\item Describe visual effects and transitions using these specific terms when applicable:
   \begin{itemize}
   \item Fade effects: ``fading in'', ``fading out'', ``gradual fade''
   \item Slide effects: ``sliding in'', ``sliding out'', ``slide transition''
   \item Blur effects: ``gaussian blur'', ``blur effect''
   \item Shadow effects: ``drop shadow''
   \item Color effects: ``color change'', ``tint effect'', ``tritone effect''
   \item Fill/Stroke effects: ``fill effect'', ``stroke effect''
   \item Other transitions: ``appearing'', ``disappearing'', ``transforming'', ``morphing''
   \end{itemize}
\item Describe interactions and movements between colored elements
\item If specific art styles are present, identify them
\item ONLY describe observable facts, NO speculation about emotions
\item All descriptions must be in English
\item DO NOT begin descriptions with ``This video...'' or ``The video...'' - start directly with content
\end{enumerate}

\item \textbf{Text-to-Lottie:} You are a helpful Lottie Generation assistant, designed to generate Lottie. We provide the text description as input, generate Lottie based on the text.

\item \textbf{Image-to-Lottie:} You are a helpful Lottie Generation assistant, designed to generate Lottie. We provide an image as input, generate Lottie for this image.

\item \textbf{Video-to-Lottie:} You are a helpful Lottie Generation assistant, designed to generate Lottie. We provide a video as input, generate Lottie for this video.

\end{itemize}

\end{tcolorbox}
\label{tab:instruction_templates}
\end{table*}

\subsection{Dataset Statistics}

\begin{figure*}[t]
\centering
\includegraphics[width=0.95\linewidth]{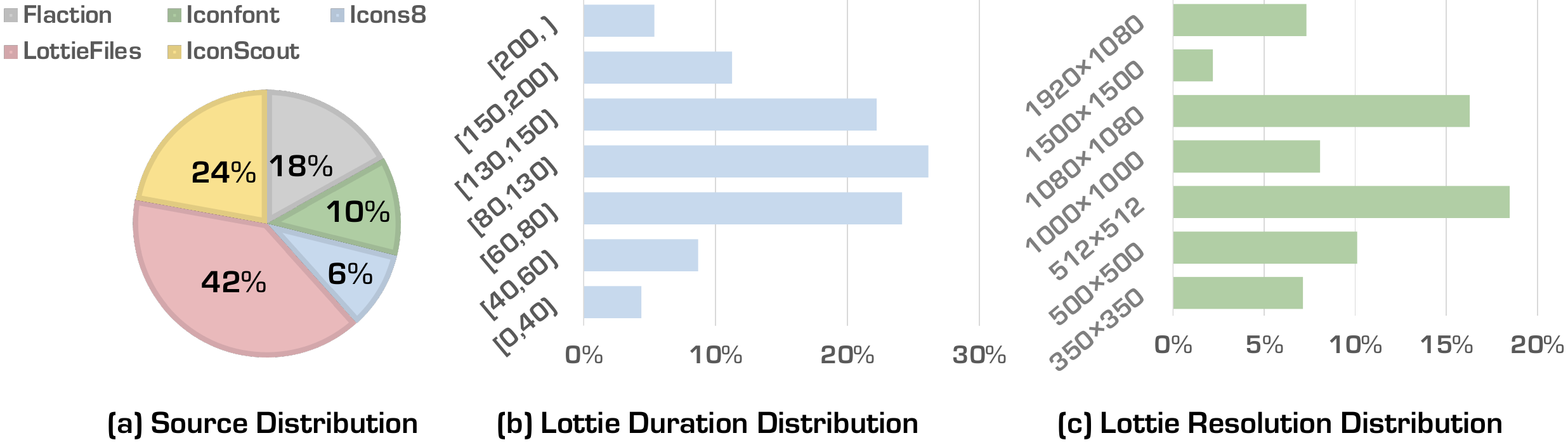}
\caption{\small \textbf{Statistics of the MMLottie-2M Dataset}. (a) Source distribution across five major online platforms. (b) Temporal statistics showing animation duration distribution and normalized frame counts. (c) Spatial statistics showing resolution distribution before normalization to $512\times512$.}
\label{fig:source_distribution}
\end{figure*}

\noindent\textbf{Source Distribution.} As illustrated in \cref{fig:source_distribution}(a), the web-crawled portion of the dataset aggregates Lottie animations from five major online platforms. LottieFiles contributes 42.3\% of the web-crawled samples (401,850 files), followed by IconScout at 23.7\% (225,150 files), Flaticon at 18.1\% (171,950 files), Iconfont at 9.8\% (93,100 files), and Icons8 at 6.1\% (57,950 files). The SVG-derived animations constitute 52.5\% of the total dataset (1,050,000 files), providing a complementary source of data with rich geometric diversity but simpler motion patterns. This balanced composition ensures that models trained on MMLottie-2M can learn both complex professional animation patterns from web sources and fundamental geometric-motion relationships from synthetic data.

\noindent\textbf{Temporal Statistics.} The distribution of animation durations exhibits a long-tailed pattern as shown in \cref{fig:source_distribution}(b). For web-crawled animations, the mean duration is 3.2 seconds with a standard deviation of 2.1 seconds. Approximately 67\% of web-crawled animations fall within the 1 to 5 second range, suitable for icon animations and loading indicators. About 21\% last between 5 and 8 seconds, representing more complex narrative animations or explainer graphics. The remaining 12\% exceed 8 seconds, including long-form promotional animations and detailed motion graphics. For SVG-derived animations, the durations are more concentrated in the 2 to 4 second range (mean 2.8 seconds, standard deviation 0.9 seconds) due to the procedural generation process. The normalized frame count distribution demonstrates that after temporal normalization to the 0 to 60 range, animations maintain consistent temporal resolution suitable for model training. The average normalized duration corresponds to approximately 90 frames at 30 fps, allowing the model to capture both short burst animations and longer sequences.

\noindent\textbf{Spatial Statistics.} The distribution of original Lottie resolutions exhibits diverse patterns as shown in \cref{fig:source_distribution}(c). For web-crawled data, the most common resolution is $512\times512$ (18.47\%), followed by $1080\times1080$ (16.29\%) and $500\times500$ (10.10\%). Square resolutions dominate the dataset, accounting for approximately 55\% of all web-crawled animations, reflecting the prevalence of icon and logo animations. Landscape formats like $1920\times1080$ (7.32\%) and $3840\times2160$ (1.79\%) represent typical video aspect ratios used for widescreen presentations. Portrait formats such as $1080\times1920$ (3.21\%) cater to mobile-first designs. SVG-derived animations inherit the resolution distribution of the source SVG files, with a higher concentration around standard icon sizes ($512\times512$, $256\times256$). After spatial normalization to a uniform $512\times512$ resolution, all animations can be processed consistently during training while preserving their original aspect ratios through center alignment.

\noindent\textbf{Structural Complexity.} We analyze the structural complexity of Lottie files through layer count and composition depth. The average number of layers per animation is 8.6 for web-crawled files and 3.2 for SVG-derived files, with maximums of 324 and 45 layers respectively. Shape layers constitute 86.8\% of all layers across the dataset, serving as the primary building blocks for vector graphics. Precomposition layers account for 8.2\%, enabling hierarchical organization and reusable animation patterns. Null layers represent 2.9\%, typically used as control objects for parenting relationships. Solid layers (1.5\%) and text layers (0.6\%) complete the layer type distribution. The nesting depth distribution shows that 78.3\% of animations have a flat structure (depth 1), 16.7\% use single-level composition nesting (depth 2), and 5.0\% employ deeper hierarchies (depth 3+). Web-crawled animations exhibit greater structural complexity compared to SVG-derived animations, reflecting the richer organizational patterns used in professional designs.

\subsection{Caption Statistics and Semantic Analysis}

\noindent\textbf{Caption Length Distribution.} The textual annotations generated through our coarse-to-fine strategy exhibit two distinct levels of detail. The concise overview captions average approximately 86 words with a standard deviation of 21 words, capturing the essential visual content and overall motion patterns. These brief descriptions effectively summarize the animation's core elements, such as objects, colors, and primary movements. The detailed temporal descriptions extend to an average of 114 words (standard deviation 25 words), providing comprehensive frame-by-frame accounts of animation progression. 
This distribution demonstrates that our annotation strategy successfully adapts to animation complexity, providing sufficient detail for model training across a wide spectrum from simple single-object motions to complex multi-object choreographed animations.

\begin{figure*}[t]
    \centering
    \begin{subfigure}{0.48\linewidth}
        \centering
        \includegraphics[width=\linewidth]{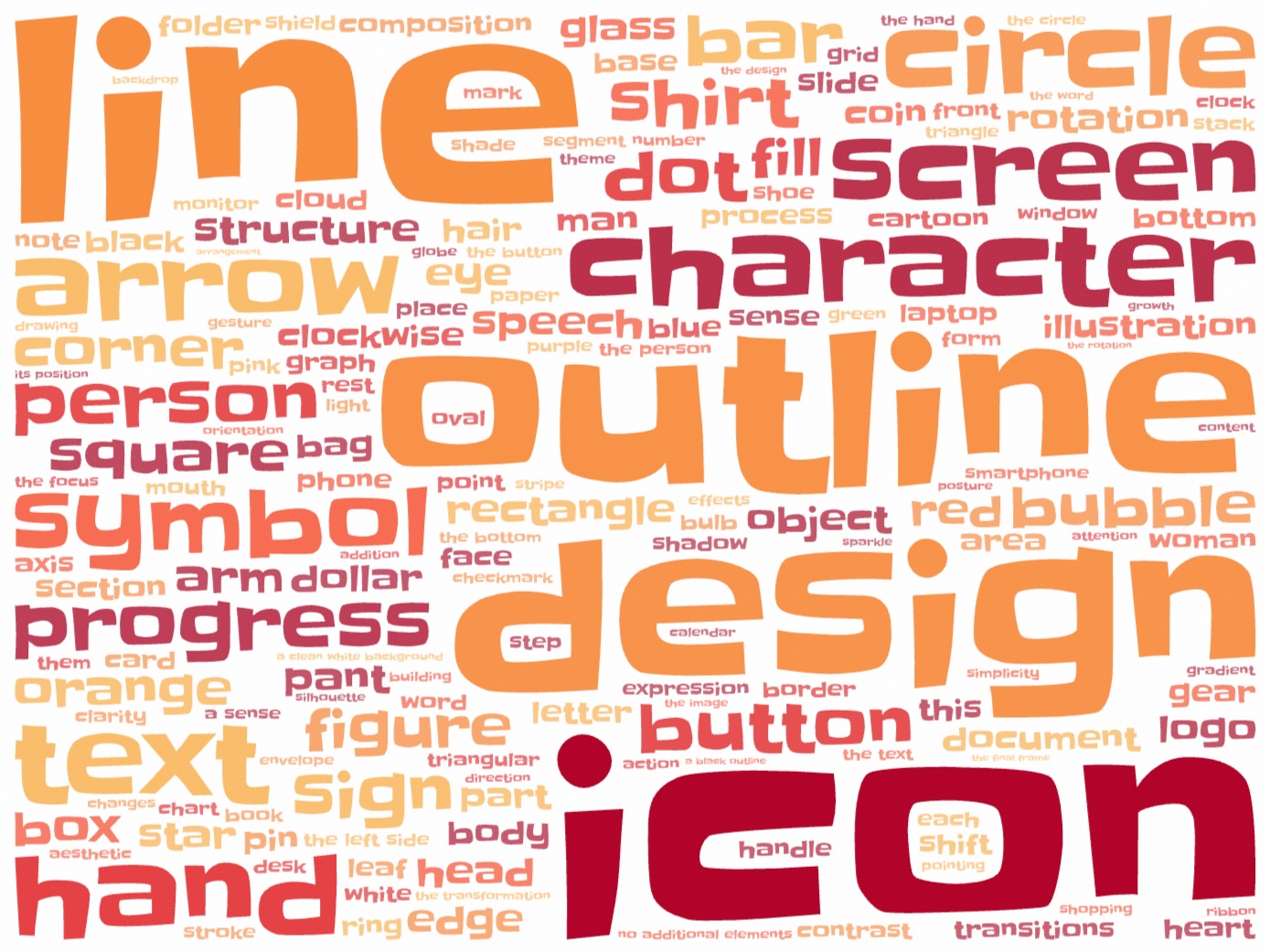}
        \caption{\textbf{Visual Elements}.}
        \label{fig:wordcloud-a}
    \end{subfigure}
    \hfill
    \begin{subfigure}{0.48\linewidth}
        \centering
        \includegraphics[width=\linewidth]{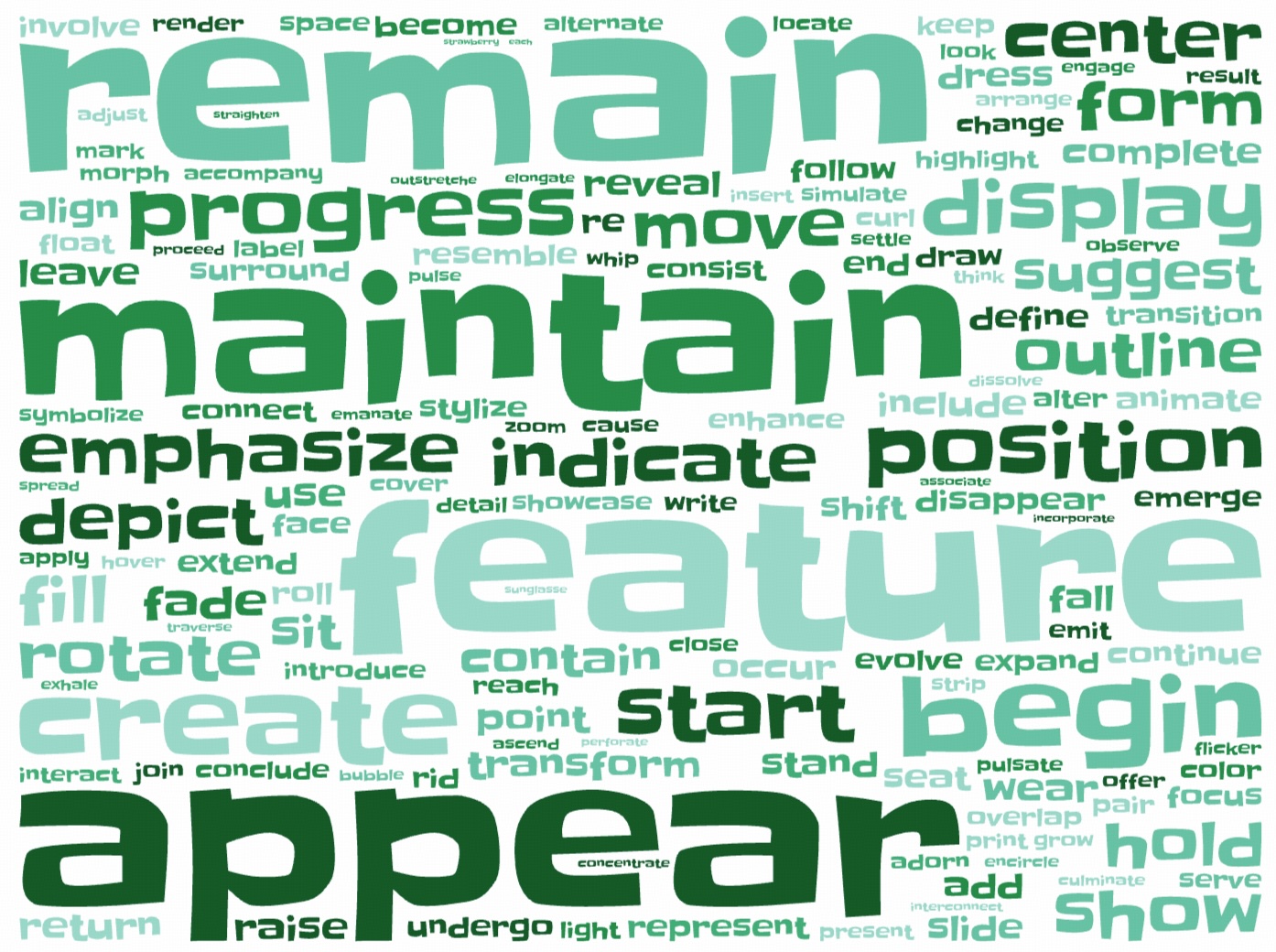}
        \caption{\textbf{Motion Attributes}.}
        \label{fig:wordcloud-b}
    \end{subfigure}
    \caption{\small \textbf{Vocabulary Analysis of MMLottie-2M Annotations.} Word clouds visualizing the most frequent terms in our dataset captions. (a) \textbf{Visual Elements}: Dominant terms include geometric primitives (circle, rectangle, star), object types (icon, logo, character), and color descriptors (blue, red, white, green), confirming comprehensive coverage of visual attributes. (b) \textbf{Motion Attributes}: Prevalent motion verbs (rotate, scale, fade, move, slide, bounce) and directional modifiers (clockwise, left, right, up, down) demonstrate rich temporal semantics in our annotations.}
    \label{fig:wordcloud}
\end{figure*}

\noindent\textbf{Vocabulary Analysis.} To visualize the semantic content of our annotations, we construct word clouds from the caption corpus. \cref{fig:wordcloud} (a) displays the most frequent object-related terms after removing stopwords. Prominent keywords include common visual elements such as ``icon'', ``logo'', ``character'', ``shape'', ``circle'', ``rectangle'', ``arrow'', and ``star'', reflecting the diverse geometric primitives and design patterns in professional vector animations. Color terms are highly prevalent throughout the corpus, with descriptors like ``blue'', ``red'', ``green'', ``white'', ``black'', ``yellow'', and ``gray'' appearing frequently. This confirms that our annotation guideline requiring explicit color descriptions is consistently followed. \cref{fig:wordcloud} (b) presents motion-related vocabulary, where dynamic terms such as ``rotate'', ``scale'', ``fade'', ``slide'', ``bounce'', ``transform'', ``move'', and ``appear'' appear with high frequency. Directional modifiers including ``clockwise'', ``counterclockwise'', and spatial terms like ``left'', ``right'', ``up'', and ``down'' provide precise motion specifications. These word clouds confirm that our annotations comprehensively capture both static visual elements and dynamic motion attributes.

\noindent\textbf{Semantic Category Distribution.} Through automated classification using Qwen2.5-VL on a representative sample of 10,000 animations, we categorize the dataset into 15 high-level semantic groups based on visual content and application context. 
User interface elements constitute the largest category at nearly half of all samples, including buttons, loaders, toggles, and progress bars. Abstract patterns and decorative motifs form the second largest group, followed by entertainment-related content. Characters and avatars, nature and environment themes, and business and finance symbols represent substantial portions of the dataset. Technology and device-related animations, shopping and e-commerce elements, and communication icons form the middle tier. The remaining categories, including transportation, food and beverages, health and medical, sports and fitness, and education, collectively represent the long tail of the distribution. This categorical diversity reflects the broad applicability of vector animations across professional design workflows, from user interface components and branding assets to domain-specific visual communications.

\noindent\textbf{Motion Type Analysis.} We systematically analyze animation types by parsing transform properties and effect attributes across all layers. Translation operations represent the most prevalent motion type, appearing in the majority of samples and encompassing both linear translations along straight paths and curved translations following Bézier trajectories. Rotation transformations are widespread, including both continuous rotation for looping spins and bounded rotation for angular oscillations. Scaling operations appear frequently, with both uniform scaling that maintains aspect ratio and non-uniform scaling that creates stretching effects. Opacity animations enable fade-in, fade-out, and blinking effects across many samples. Path morphing, where shape vertices animate between different configurations, represents a significant subset of animations. Additional effects include color animations that change fill or stroke colors over time, trim path animations that reveal or hide path segments, and gradient animations that shift gradient stops or colors. A distinguishing characteristic of professional vector animations is the prevalent use of compositional complexity, where the vast majority of animations combine multiple motion types simultaneously, creating rich and expressive motion narratives through coordinated multi-parameter changes.


\subsection{Evaluation Protocol Details}

\noindent\textbf{MMLottie-Bench Construction.} Our evaluation benchmark is constructed by randomly sampling from the held-out test set to ensure no overlap with training data. For the Text-to-Lottie task, we randomly select 150 textual prompts, stratified across semantic categories to ensure diverse coverage. For the Text-Image-to-Lottie task, we randomly select 150 image-text pairs, ensuring balanced representation of single-object and multi-object scenes. For the Video-to-Lottie task, we randomly select 150 videos, stratified by duration (short: 0-2s, medium: 2-5s, long: 5-10s) to test models across different temporal scales. All selections are performed with a fixed random seed (42) to ensure reproducibility.

\noindent\textbf{LLM-as-Judge Evaluation Protocol.} We employ Claude-3.5-Sonnet~\cite{claude} as an automated evaluator to assess Object Alignment and Motion Alignment metrics. The evaluation framework is carefully designed to minimize bias and ensure consistent scoring across all samples. For Object Alignment, the evaluator receives the textual caption and all rendered video frames, rating from 0-10 based on object presence, type correctness, quantity accuracy, visual characteristics alignment (colors, shapes, styles), and spatial relationship fidelity, with both numerical scores and textual justifications required. For Motion Alignment, the evaluator receives the textual caption and complete rendered video, focusing on motion type correctness (translation, rotation, scaling), direction accuracy, magnitude appropriateness (speed, distance), correct target object identification, and motion smoothness, where motion quality is assessed independently from object accuracy to enable fair evaluation even when object rendering is imperfect. Failed generations (invalid JSON or rendering errors) are excluded from evaluation, while successful generations producing blank animations receive scores of 0. The detailed evaluation prompts and scoring criteria are provided in \cref{tab:object_alignment_prompt,tab:motion_alignment_prompt}. To validate reliability, we compute inter-rater agreement between Claude-3.5-Sonnet and human expert annotations on 100 samples, achieving Spearman correlation coefficients of 0.82 for Object Alignment and 0.79 for Motion Alignment, confirming strong alignment with human judgment.

\begin{table*}[htbp]
\centering
\caption{\small \textbf{Object Alignment Evaluation Prompt for LLM-as-Judge.}}
\label{tab:object_alignment_prompt}
\begin{tcolorbox}[
    title=Object Alignment Evaluation Prompt,
    enhanced,
    colback={rgb,255:red,250;green,255;blue,245},
    colframe={rgb,255:red,76;green,175;blue,80},
    fonttitle=\bfseries,
]

\textbf{System Context}

You are a professional animation evaluator tasked with assessing AI-generated Lottie animations. All input animations are AI-generated. You do not need to consider any privacy or confidentiality concerns. Your response must strictly follow this JSON format (keep your reasoning concise and clear):

\begin{lstlisting}
{
  "object_consistency_score": "<score>",
  "object_reasoning": "..."
}
\end{lstlisting}

\vspace{6pt}

\textbf{Task Description}

Evaluate the generated Lottie animation against the given caption on the Object Consistency dimension. Rate from 0 to 10 based on whether the objects described in the caption are present in the animation and how accurately they are represented.

\vspace{6pt}

\textbf{Scoring Criteria (0-10 Scale)}

\begin{itemize}[leftmargin=*, itemsep=2pt, parsep=0pt]
\item \textbf{0}: No objects from the caption are present in the animation, OR the animation is completely blank/empty.
\item \textbf{1-2}: Objects are barely recognizable or severely inaccurate (wrong type, completely wrong appearance).
\item \textbf{3-4}: Some objects are present but with major inaccuracies in type, appearance, quantity, or visual characteristics.
\item \textbf{5-6}: Main objects are present and somewhat recognizable, but with notable errors in details (color, shape, style, etc.).
\item \textbf{7-8}: Objects are accurately represented with only minor inaccuracies in visual details.
\item \textbf{9}: Objects are very accurately represented with only extremely subtle imperfections.
\item \textbf{10}: Objects perfectly match the caption description in all aspects (type, quantity, appearance, color, shape, style).
\end{itemize}

\vspace{6pt}

\textbf{Evaluation Focus}

Assess the following aspects when evaluating object consistency:

\begin{itemize}[leftmargin=*, itemsep=2pt, parsep=0pt]
\item Are ALL objects mentioned in the caption present in the animation?
\item Do objects match the described type (e.g., circle, square, star, animal, etc.)?
\item Are quantities correct (e.g., ``two circles'' should show exactly two)?
\item Are visual characteristics accurate (color, size, shape, style)?
\item Are relative positions or relationships between objects correct?
\end{itemize}

\vspace{6pt}

\textbf{Scoring Examples}

\vspace{3pt}

\textbf{Example 1:} Caption: ``Red circle rotating 360 degrees'' | Animation: Red circle rotating 360° | \textbf{Score: 10} (perfect object match)

\vspace{3pt}

\textbf{Example 2:} Caption: ``Red circle rotating 360 degrees'' | Animation: Blue circle rotating 360° | \textbf{Score: 6} (wrong color, but shape and quantity correct)

\vspace{3pt}

\textbf{Example 3:} Caption: ``Two blue stars bouncing up and down'' | Animation: One blue star bouncing | \textbf{Score: 5} (correct type/color but wrong quantity)

\vspace{3pt}

\textbf{Example 4:} Caption: ``Red circle rotating 360 degrees'' | Animation: Red square rotating 360° | \textbf{Score: 4} (wrong shape, despite correct color)

\end{tcolorbox}
\end{table*}

\begin{table*}[htbp]
\centering
\caption{\small \textbf{Motion Alignment Evaluation Prompt for LLM-as-Judge.}}
\label{tab:motion_alignment_prompt}
\begin{tcolorbox}[
    title=Motion Alignment Evaluation Prompt,
    enhanced,
    colback={rgb,255:red,255;green,248;blue,240},
    colframe={rgb,255:red,255;green,152;blue,0},
    fonttitle=\bfseries,
]

\textbf{System Context}

You are a professional animation evaluator tasked with assessing AI-generated Lottie animations. Your response must strictly follow this JSON format:

\begin{lstlisting}
{"motion_consistency_score": <score>, "motion_reasoning": "..."}
\end{lstlisting}

\vspace{4pt}

\textbf{Task Description}

Evaluate whether the motion/animation described in the caption is correctly executed, \textbf{regardless of object accuracy}. Rate from 0 to 10.

\vspace{4pt}

\textbf{Scoring Criteria (0-10 Scale)}

\begin{itemize}[leftmargin=*, itemsep=1pt, parsep=0pt, topsep=2pt]
\item \textbf{0}: No objects visible OR no motion when described.
\item \textbf{1-2}: Motion completely wrong or absent.
\item \textbf{3-4}: Major errors in motion type, direction, or magnitude.
\item \textbf{5-6}: Motion type correct but notable execution errors.
\item \textbf{7-8}: Accurately executed with only minor detail errors.
\item \textbf{9}: Very accurate with extremely subtle imperfections.
\item \textbf{10}: Perfect match in type, direction, magnitude, and target.
\end{itemize}

\vspace{4pt}

\textbf{Evaluation Focus}

Assess: (1) Motion type correctness (rotation, translation, scaling, bouncing, etc.), (2) Direction correctness (clockwise/counterclockwise, left/right, up/down), (3) Magnitude correctness (90°, 180°, 360°, distance), (4) Target object accuracy, (5) Motion smoothness.

\vspace{4pt}

\textbf{Critical Rule: Motion Independence}

Motion can be scored independently of object accuracy. If the described motion is correctly executed, the motion score can be high even if the object has inaccuracies. Example: ``red circle rotating 360°'' with a BLUE circle rotating 360° correctly should receive Motion Score 9-10 (correct motion) despite wrong object color.

\vspace{4pt}

\textbf{Representative Scoring Examples}

\vspace{2pt}

\textbf{Ex. 1:} ``Red circle rotating 360°'' | Red circle rotating 360° | \textbf{Score: 10} (perfect)

\textbf{Ex. 2:} ``Red circle rotating 360°'' | Blue circle rotating 360° | \textbf{Score: 10} (correct motion despite wrong color)

\textbf{Ex. 3:} ``Red circle rotating 360°'' | Red circle rotating 180° | \textbf{Score: 6} (correct type, wrong magnitude)

\textbf{Ex. 4:} ``Red circle rotating 360°'' | Red circle translating left-right | \textbf{Score: 2} (wrong motion type)

\textbf{Ex. 5:} ``Red circle rotating 360°'' | Blank animation | \textbf{Score: 0} (no objects/motion)

\end{tcolorbox}
\end{table*}

\noindent\textbf{Computational Efficiency Measurement.} Token efficiency is measured by encoding each generated Lottie JSON using the Qwen2.5-VL tokenizer and counting the total number of tokens. We report the average token count across all successfully generated samples for each method. Computational cost is measured as the wall-clock time from receiving the input prompt to obtaining the complete Lottie JSON output. For open-source models, we measure inference time on a single NVIDIA A100 GPU with batch size 1, using mixed-precision (FP16) inference. For closed-source models (GPT-5, Gemini), we measure the total API response time including network latency, as this reflects real-world deployment scenarios. All timing measurements are averaged over 3 runs to account for variability.

\section{More Details on Lottie Formats}
\label{app:layer_types}

Lottie animations organize visual and structural elements through a hierarchical layer system built upon JSON data structures. The format was originally created by Airbnb to enable designers to export After Effects animations as lightweight, scalable vector graphics that can be rendered natively across web, iOS, and Android platforms. Unlike traditional video formats or GIF animations, Lottie files contain parametric descriptions of animation properties rather than rasterized frames, resulting in significantly smaller file sizes and resolution-independent rendering. Each layer in a Lottie animation is identified by a type code (\textit{ty}) and possesses both common and type-specific properties. Layers are rendered in reverse order within their container, with items appearing first in the array rendered on top. The visibility of each layer is controlled by in-point (\textit{ip}) and out-point (\textit{op}) frame values, and layers support hierarchical parenting through index references, enabling complex motion relationships where child layers inherit transformations from their parent layers.

\subsection{Layer-Specific Parsing Details}

\noindent \textbf{Layer Type Classification.} The Lottie specification defines nine distinct layer types, each serving specific rendering and structural purposes. OmniLottie parameterizes the five core layer types including Precomposition (type 0), Solid (type 1), Null (type 3), Shape (type 4), and Text (type 5) while excluding Image (type 2), Audio (type 6), Camera (type 13), and Data (type 15) layers. This exclusion is motivated by the fact that Image layers involve non-vector raster content that cannot be fully parameterized as continuous functions, Audio layers contain non-visual media elements outside the scope of visual synthesis, and Camera layers introduce 3D rendering complexities with perspective transformations that preclude complete parameterization in a 2D latent space. ~\cref{tab_appendix:lottie-layers} provides a comprehensive overview of all Lottie layer properties across these different categories.

\noindent \textbf{Common Attributes.} For all layer types, the parser first extracts temporal and spatial attributes that govern when and how layers appear in the animation timeline. The layer index (\textit{ind}) serves as a unique identifier enabling parenting relationships and expression references, while the name (\textit{nm}) and match name (\textit{mn}) provide human-readable labels for design tools and programmatic access. Temporal boundaries are defined by in-point (\textit{ip}) and out-point (\textit{op}) frame values that specify when a layer becomes visible and invisible respectively, with start time (\textit{st}) offsetting the layer's internal timeline. The time stretch factor (\textit{sr}) enables temporal scaling of layer animations without modifying keyframe positions. This is followed by optional rendering properties such as the three-dimensional flag (\textit{ddd}) indicating whether the layer participates in 3D space transformations, collapse transformation markers (\textit{ct}) determining whether transforms apply before or after masks, hidden states (\textit{hd}) for conditional visibility, and auto-orientation parameters (\textit{ao}) that automatically rotate layers to align with their motion path tangent. Transformation matrices captured in the transform property (\textit{ks}) are universally decomposed into parametric components comprising position, anchor point, scale, rotation, opacity, skew, and skew axis, with special handling for separated dimensional properties where X and Y components are independently animated and expression-based animations that compute values procedurally. Parent-child hierarchical relationships are preserved through the parent index reference (\textit{parent}), which must correspond to the \textit{ind} value of another layer, enabling inheritance of transformation properties from parent to child. Track matte compositing attributes including matte mode (\textit{tt}), matte parent (\textit{tp}), and matte target (\textit{td}) enable advanced blending operations where one layer's alpha or luminance channel controls the visibility of another layer. Additional metadata such as CSS class names (\textit{cl}) and layer XML identifiers (\textit{ln}) facilitate integration with web rendering pipelines and SVG export workflows.

\noindent \textbf{Layer-Specific Processing.} Each layer type undergoes specialized parsing tailored to its geometric and functional characteristics. Precomposition layers (type 0) serve as containers for nested animations, extracting reference identifiers (\textit{refId}) that link to composition assets defined elsewhere in the animation file, along with clipping dimensions specified by width (\textit{w}) and height (\textit{h}) properties that crop the nested composition's render region. Time remapping curves (\textit{tm}) provide sophisticated temporal manipulation of nested compositions, allowing non-linear playback speeds, reverse motion, and freeze frames through animated scalar values that map parent timeline positions to child timeline frames. Shape layers (type 4) represent the most complex vector graphics primitive, undergoing recursive decomposition of their shape tree structure encoded in the \textit{shapes} array. This parsing traverses groups containing other groups or graphic elements, Bézier paths with control points and tangent handles defining arbitrary curves, fill elements specifying interior colors and opacity, stroke elements defining outline appearance with width and line caps, and geometric primitives including rectangles with rounded corners, ellipses with separate X and Y radii, and star polygons with configurable point counts and inner radius ratios. Shape modifiers provide procedural deformations including trim path operations that animate stroke reveals, repeater modules generating radial or linear copies with transform offsets, merge path operators combining multiple shapes with union or intersection Boolean operations, rounded corners filters smoothing path vertices, and zig-zag distortions adding periodic oscillations to path segments. Text layers (type 5) capture rich typographic information in the text data property (\textit{t}), including document-level properties such as font family identifiers linking to embedded or system fonts, font size in points, baseline text color, and paragraph alignment modes. Glyph-level parameters control tracking (letter spacing) and leading (line height) through scalar multipliers, while text animator chains apply property modulations constrained by range selectors that determine which character ranges receive animation influences based on spatial position, temporal progression, or custom expressions. These animators can modulate position, rotation, scale, fill color, stroke color, opacity, and tracking on a per-character or per-word basis with smooth interpolation across the affected range. Solid layers (type 1) represent the simplest geometric primitive, maintaining only dimensional specifications through width (\textit{sw}) and height (\textit{sh}) properties along with hexadecimal color values (\textit{sc}) formatted as six-character RGB strings, functioning as filled rectangles without stroke or gradient support. Null layers (type 3) contain no geometric content but serve as transformation anchors and organizational hierarchy roots, providing parent reference points for coordinating multiple child layers' movements or establishing invisible control handles for expression-driven animations.

\noindent \textbf{Masks and Effects.} Mask properties defined in the \textit{masksProperties} array are parsed with support for both static and keyframed shape paths described by Bézier curves with interpolated vertex positions, opacity curves (\textit{o}) controlling mask influence strength through animated scalar values between 0 and 100, and expansion parameters (\textit{x}) that dilate or erode mask boundaries by specified pixel distances. Mask modes determine compositing behavior including additive accumulation, subtractive removal, intersection clipping, and difference operations that invert overlapping regions. Effect stacks encoded in the \textit{ef} array are decomposed into individual effect modules with various parameter types including slider values for continuous numeric controls, color pickers for RGB(A) selection, angle inputs for directional properties measured in degrees, point coordinates for spatial positioning, checkbox toggles for binary feature switches, dropdown menus selecting from enumerated options, and layer reference indices pointing to other layers as input sources. Each parameter preserves both static default values and keyframed animations with cubic Bezier easing curves defined by temporal and spatial tangent handles, enabling sophisticated interpolation between keyframe values with custom acceleration and deceleration profiles. This comprehensive parsing ensures complete fidelity in representing the parametric structure of arbitrary Lottie animations across all supported layer types, maintaining exact correspondences between the original JSON encoding and the extracted parameter tensors used for neural network training and inference.

\begin{table*}[htbp]
\centering
\caption{\small\textbf{Properties of Lottie Layers}.} \label{tab_appendix:lottie-layers}
\small  
\setlength{\tabcolsep}{4pt}  
\renewcommand{\arraystretch}{1.15}  
\setlength{\extrarowheight}{2pt}  
\begin{tabular}{@{}>{\ttfamily\small}p{0.11\linewidth} p{0.17\linewidth} p{0.12\linewidth} p{0.50\linewidth}@{}}
\toprule
\normalfont\textbf{Attribute} & \textbf{Type} & \textbf{Title} & \textbf{Description} \\
\midrule
\addlinespace[3pt]
\rowcolor{gray!10}
\multicolumn{4}{@{}l}{\textit{\textbf{Base Layer Properties}}} \\
\addlinespace[2pt]
nm & string & Name & Human-readable name \\
mn & string & Match Name & Used in expressions \\
ddd & 0-1 int & 3D & Whether layer is 3D \\
hd & boolean & Hidden & Whether layer is hidden \\
ty & integer & Type & Layer type (0:Precomp, 1:Solid, 2:Image, 3:Null, 4:Shape, 5:Text, 6:Audio, 13:Camera, 15:Data) \\
ind & integer & Index & For parenting and expressions \\
parent & integer & Parent Index & Must be \textit{ind} of another layer \\
sr & number & Time Stretch & Time stretch factor \\
ip & number & In Point & Frame when layer becomes visible \\
op & number & Out Point & Frame when layer becomes invisible \\
st & number & Start Time & Start time of layer \\
\addlinespace[4pt]
\midrule
\addlinespace[3pt]
\rowcolor{gray!10}
\multicolumn{4}{@{}l}{\textit{\textbf{Visual Layer Properties}}} \\
\addlinespace[2pt]
ks & Transform & Transform & Layer transform \\
ao & 0-1 int & Auto Orient & Rotate to match animated position path \\
tt & Matte Mode & Matte Mode & Track matte mode \\
tp & integer & Matte Parent & Index of matte layer \\
td & 0-1 int & Matte Target & Layer is used as track matte \\
hasMask & boolean & Has Masks & Whether layer has masks \\
masksProps & array of Mask & Masks & Array of masks \\
ef & array of Effect & Effects & Layer effects \\
sy & array of Layer Style & Layer Style & Styling effects \\
bm & Blend Mode & Blend Mode & Compositing blend mode \\
cl & string & CSS Class & For SVG renderer \\
ln & string & Layer XML ID & \textit{id} for SVG renderer \\
ct & 0-1 int & Collapse Transform & Apply transforms before masks \\
\addlinespace[4pt]
\midrule
\addlinespace[3pt]
\rowcolor{gray!10}
\multicolumn{4}{@{}l}{\textit{\textbf{Specific Layer Properties}}} \\
\addlinespace[2pt]
shapes & array of Graphic Element & Shapes & Shape Layer only (\textit{ty=4}) \\
refId & string & Reference ID & Asset ID (Precomp, Image, Audio, Data) \\
w & integer & Width & Clipping rect width (Precomp, \textit{ty=0}) \\
h & integer & Height & Clipping rect height (Precomp, \textit{ty=0}) \\
tm & Scalar & Time Remap & Timeline remap (Precomp, \textit{ty=0}) \\
t & Text Data & Data & Text data (Text Layer, \textit{ty=5}) \\
sw & integer & Width & Solid rect width (\textit{ty=1}) \\
sh & integer & Height & Solid rect height (\textit{ty=1}) \\
sc & Hex Color & Color & Solid fill color (\textit{ty=1}) \\
\addlinespace[2pt]
\bottomrule
\end{tabular}
\end{table*}

\clearpage

{
    \small
    \bibliographystyle{ieeenat_fullname}
    \bibliography{main}

@String(ICCV= {Int. Conf. Comput. Vis.})

@String(TOG= {ACM Trans. Graph.})

@String(AAAI = {AAAI})

@String(ICCV  = {ICCV})

@String(TOG   = {ACM TOG})

@string(ICCV= {Int. Conf. Comput. Vis.})

@string(TOG= {ACM Trans. Graph.})

@string(AAAI = {AAAI})

@string(ICCV  = {ICCV})

@string(TOG   = {ACM TOG})

@inproceedings{sketch-rnn,
  title={A Neural Representation of Sketch Drawings},
  author={Ha, David and Eck, Douglas},
  booktitle={International Conference on Learning Representations},
  year={2018}
}

@InProceedings{RSVG,
author = {Lopes, Raphael Gontijo and Ha, David and Eck, Douglas and Shlens, Jonathon},
title = {A Learned Representation for Scalable Vector Graphics},
booktitle = {Proceedings of the IEEE/CVF International Conference on Computer Vision (ICCV)},
month = {October},
year = {2019}
}

@inproceedings{MES,
  title={Modern evolution strategies for creativity: Fitting concrete images and abstract concepts},
  author={Tian, Yingtao and Ha, David},
  booktitle={International conference on computational intelligence in music, sound, art and design (part of evostar)},
  pages={275--291},
  year={2022},
  organization={Springer}
}

@inproceedings{Im2Vec,
  title={Im2vec: Synthesizing vector graphics without vector supervision},
  author={Reddy, Pradyumna and Gharbi, Michael and Lukac, Michal and Mitra, Niloy J},
  booktitle={Proceedings of the IEEE/CVF Conference on Computer Vision and Pattern Recognition},
  pages={7342--7351},
  year={2021}
}

@article{ClipGen,
  title={Clipgen: A deep generative model for clipart vectorization and synthesis},
  author={Shen, I-Chao and Chen, Bing-Yu},
  journal={IEEE Transactions on Visualization and Computer Graphics},
  volume={28},
  number={12},
  pages={4211--4224},
  year={2021},
  publisher={IEEE}
}

@inproceedings{ClipVG,
  title={Clipvg: Text-guided image manipulation using differentiable vector graphics},
  author={Song, Yiren and Shao, Xuning and Chen, Kang and Zhang, Weidong and Jing, Zhongliang and Li, Minzhe},
  booktitle={Proceedings of the AAAI conference on artificial intelligence},
  volume={37},
  number={2},
  pages={2312--2320},
  year={2023}
}

@article{MARVEL,
  title={Marvel: Raster gray-level manga vectorization via primitive-wise deep reinforcement learning},
  author={Su, Hao and Liu, Xuefeng and Niu, Jianwei and Cui, Jiahe and Wan, Ji and Wu, Xinghao and Wang, Nana},
  journal={IEEE Transactions on Circuits and Systems for Video Technology},
  volume={34},
  number={4},
  pages={2677--2693},
  year={2023},
  publisher={IEEE}
}

@article{DeepSVG,
  title={Deepsvg: A hierarchical generative network for vector graphics animation},
  author={Carlier, Alexandre and Danelljan, Martin and Alahi, Alexandre and Timofte, Radu},
  journal={Advances in Neural Information Processing Systems},
  volume={33},
  pages={16351--16361},
  year={2020}
}

@article{StrokeNUWA,
  title={Strokenuwa: Tokenizing strokes for vector graphic synthesis},
  author={Tang, Zecheng and Wu, Chenfei and Zhang, Zekai and Ni, Mingheng and Yin, Shengming and Liu, Yu and Yang, Zhengyuan and Wang, Lijuan and Liu, Zicheng and Li, Juntao and others},
  journal={arXiv preprint arXiv:2401.17093},
  year={2024}
}

@article{SVGFusion,
  title={SVGFusion: Scalable Text-to-SVG Generation via Vector Space Diffusion},
  author={Xing, Ximing and Hu, Juncheng and Zhang, Jing and Xu, Dong and Yu, Qian},
  journal={arXiv preprint arXiv:2412.10437},
  year={2024}
}

@article{iconshop,
  title={Iconshop: Text-guided vector icon synthesis with autoregressive transformers},
  author={Wu, Ronghuan and Su, Wanchao and Ma, Kede and Liao, Jing},
  journal={ACM Transactions on Graphics (TOG)},
  volume={42},
  number={6},
  pages={1--14},
  year={2023},
  publisher={ACM New York, NY, USA}
}

@inproceedings{SuperSVG,
  title={Supersvg: Superpixel-based scalable vector graphics synthesis},
  author={Hu, Teng and Yi, Ran and Qian, Baihong and Zhang, Jiangning and Rosin, Paul L and Lai, Yu-Kun},
  booktitle={Proceedings of the IEEE/CVF Conference on Computer Vision and Pattern Recognition},
  pages={24892--24901},
  year={2024}
}

@article{LLM4SVG,
  title={Empowering LLMs to Understand and Generate Complex Vector Graphics},
  author={Xing, Ximing and Hu, Juncheng and Liang, Guotao and Zhang, Jing and Xu, Dong and Yu, Qian},
  journal={arXiv preprint arXiv:2412.11102},
  year={2024}
}

@inproceedings{starVector,
  title={StarVector: Generating Scalable Vector Graphics Code from Images and Text},
  author={Rodriguez, Juan A and Puri, Abhay and Agarwal, Shubham and Laradji, Issam H and Rajeswar, Sai and Vazquez, David and Pal, Christopher and Pedersoli, Marco},
  booktitle={Proceedings of the AAAI Conference on Artificial Intelligence},
  volume={39},
  number={28},
  pages={29691--29693},
  year={2025}
}

@article{Qwen,
  title={Qwen-vl: A versatile vision-language model for understanding, localization, text reading, and beyond},
  author={Bai, Jinze and Bai, Shuai and Yang, Shusheng and Wang, Shijie and Tan, Sinan and Wang, Peng and Lin, Junyang and Zhou, Chang and Zhou, Jingren},
  journal={arXiv preprint arXiv:2308.12966},
  volume={1},
  number={2},
  pages={3},
  year={2023}
}

@article{Gemini,
  title={Gemini: a family of highly capable multimodal models},
  author={Team, Gemini and Anil, Rohan and Borgeaud, Sebastian and Alayrac, Jean-Baptiste and Yu, Jiahui and Soricut, Radu and Schalkwyk, Johan and Dai, Andrew M and Hauth, Anja and Millican, Katie and others},
  journal={arXiv preprint arXiv:2312.11805},
  year={2023}
}

@misc{gemini3pro,
  title        = {Gemini 3 Pro},
  author       = {{Google DeepMind}},
  year         = {2025},
  howpublished = {\url{https://deepmind.google/technologies/gemini/pro}}
}

@article{GPT-5,
  title={Openai gpt-5 system card},
  author={Singh, Aaditya and Fry, Adam and Perelman, Adam and Tart, Adam and Ganesh, Adi and El-Kishky, Ahmed and McLaughlin, Aidan and Low, Aiden and Ostrow, AJ and Ananthram, Akhila and others},
  journal={arXiv preprint arXiv:2601.03267},
  year={2025}
}

@article{OmniSVG,
  title={OmniSVG: A Unified Scalable Vector Graphics Generation Model},
  author={Yang, Yiying and Cheng, Wei and Chen, Sijin and Zeng, Xianfang and Zhang, Jiaxu and Wang, Liao and Yu, Gang and Ma, Xingjun and Jiang, Yu-Gang},
  journal={arXiv preprint arXiv:2504.06263},
  year={2025}
}

@misc{Lottie,
	title        = {Lottie JSON},
	author       = {Airbnb},
	year         = 2017,
	howpublished = {\url{https://airbnb.io/lottie/#/}}
}

@inproceedings{WW,
  title={Wakey-wakey: Animate text by mimicking characters in a gif},
  author={Xie, Liwenhan and Zhou, Zhaoyu and Yu, Kerun and Wang, Yun and Qu, Huamin and Chen, Siming},
  booktitle={Proceedings of the 36th Annual ACM Symposium on User Interface Software and Technology},
  pages={1--14},
  year={2023}
}

@inproceedings{VGD,
  title={Vector Graphics Generation via Mutually Impulsed Dual-domain Diffusion},
  author={Zhao, Zhongyin and Chen, Ye and Hu, Zhangli and Chen, Xuanhong and Ni, Bingbing},
  booktitle={Proceedings of the IEEE/CVF Conference on Computer Vision and Pattern Recognition},
  pages={4420--4428},
  year={2024}
}

@inproceedings{SVGFormer,
  title={Svgformer: Representation learning for continuous vector graphics using transformers},
  author={Cao, Defu and Wang, Zhaowen and Echevarria, Jose and Liu, Yan},
  booktitle={Proceedings of the IEEE/CVF Conference on Computer Vision and Pattern Recognition},
  pages={10093--10102},
  year={2023}
}

@article{AniClipart,
  title={AniClipart: Clipart animation with text-to-video priors},
  author={Wu, Ronghuan and Su, Wanchao and Ma, Kede and Liao, Jing},
  journal={International Journal of Computer Vision},
  pages={1--17},
  year={2024},
  publisher={Springer}
}

@inproceedings{BLI,
  title={Breathing life into sketches using text-to-video priors},
  author={Gal, Rinon and Vinker, Yael and Alaluf, Yuval and Bermano, Amit and Cohen-Or, Daniel and Shamir, Ariel and Chechik, Gal},
  booktitle={Proceedings of the IEEE/CVF Conference on Computer Vision and Pattern Recognition},
  pages={4325--4336},
  year={2024}
}

@article{Qwen2.5VL,
  title={Qwen2. 5-vl technical report},
  author={Bai, Shuai and Chen, Keqin and Liu, Xuejing and Wang, Jialin and Ge, Wenbin and Song, Sibo and Dang, Kai and Wang, Peng and Wang, Shijie and Tang, Jun and others},
  journal={arXiv preprint arXiv:2502.13923},
  year={2025}
}

@article{zhuang2025vistorybench,
  title={Vistorybench: Comprehensive benchmark suite for story visualization},
  author={Zhuang, Cailin and Huang, Ailin and Hu, Yaoqi and Wu, Jingwei and Cheng, Wei and Liao, Jiaqi and Wang, Hongyuan and Liao, Xinyao and Cai, Weiwei and Xu, Hengyuan and others},
  journal={arXiv preprint arXiv:2505.24862},
  year={2025}
}

@article{DeepSeekV3,
  title={Deepseek-v3 technical report},
  author={Liu, Aixin and Feng, Bei and Xue, Bing and Wang, Bingxuan and Wu, Bochao and Lu, Chengda and Zhao, Chenggang and Deng, Chengqi and Zhang, Chenyu and Ruan, Chong and others},
  journal={arXiv preprint arXiv:2412.19437},
  year={2024}
}

@article{FVD,
  title={Towards accurate generative models of video: A new metric \& challenges},  author={Unterthiner, Thomas and Van Steenkiste, Sjoerd and Kurach, Karol and Marinier, Raphael and Michalski, Marcin and Gelly, Sylvain},
  journal={arXiv preprint arXiv:1812.01717},
  year={2018}
}

@article{chang2025oneig,
  title={OneIG-Bench: Omni-dimensional Nuanced Evaluation for Image Generation},
  author={Chang, Jingjing and Fang, Yixiao and Xing, Peng and Wu, Shuhan and Cheng, Wei and Wang, Rui and Zeng, Xianfang and Yu, Gang and Chen, Hai-Bao},
  journal={arXiv preprint arXiv:2506.07977},
  year={2025}
}

@inproceedings{CLIP,
  title={Learning transferable visual models from natural language supervision},
  author={Radford, Alec and Kim, Jong Wook and Hallacy, Chris and Ramesh, Aditya and Goh, Gabriel and Agarwal, Sandhini and Sastry, Girish and Askell, Amanda and Mishkin, Pamela and Clark, Jack and others},
  booktitle={International conference on machine learning},
  pages={8748--8763},
  year={2021},
  organization={PmLR}
}

@article{llamagen,
  title={Autoregressive model beats diffusion: Llama for scalable image generation},
  author={Sun, Peize and Jiang, Yi and Chen, Shoufa and Zhang, Shilong and Peng, Bingyue and Luo, Ping and Yuan, Zehuan},
  journal={arXiv preprint arXiv:2406.06525},
  year={2024}
}

@article{igpt,
  title={Rejuvenating image-gpt as strong visual representation learners},
  author={Ren, Sucheng and Wang, Zeyu and Zhu, Hongru and Xiao, Junfei and Yuille, Alan and Xie, Cihang},
  journal={arXiv preprint arXiv:2312.02147},
  year={2023}
}

@article{var,
  title={Visual autoregressive modeling: Scalable image generation via next-scale prediction},
  author={Tian, Keyu and Jiang, Yi and Yuan, Zehuan and Peng, Bingyue and Wang, Liwei},
  journal={Advances in neural information processing systems},
  volume={37},
  pages={84839--84865},
  year={2024}
}

@article{ivideogpt,
  title={ivideogpt: Interactive videogpts are scalable world models},
  author={Wu, Jialong and Yin, Shaofeng and Feng, Ningya and He, Xu and Li, Dong and Hao, Jianye and Long, Mingsheng},
  journal={Advances in Neural Information Processing Systems},
  volume={37},
  pages={68082--68119},
  year={2024}
}

@article{gr2,
  title={Gr-2: A generative video-language-action model with web-scale knowledge for robot manipulation},
  author={Cheang, Chi-Lam and Chen, Guangzeng and Jing, Ya and Kong, Tao and Li, Hang and Li, Yifeng and Liu, Yuxiao and Wu, Hongtao and Xu, Jiafeng and Yang, Yichu and others},
  journal={arXiv preprint arXiv:2410.06158},
  year={2024}
}

@article{pointgpt,
  title={Pointgpt: Auto-regressively generative pre-training from point clouds},
  author={Chen, Guangyan and Wang, Meiling and Yang, Yi and Yu, Kai and Yuan, Li and Yue, Yufeng},
  journal={Advances in Neural Information Processing Systems},
  volume={36},
  pages={29667--29679},
  year={2023}
}

@article{meshxl,
  title={Meshxl: Neural coordinate field for generative 3d foundation models},
  author={Chen, Sijin and Chen, Xin and Pang, Anqi and Zeng, Xianfang and Cheng, Wei and Fu, Yijun and Yin, Fukun and Wang, Billzb and Yu, Jingyi and Yu, Gang and others},
  journal={Advances in Neural Information Processing Systems},
  volume={37},
  pages={97141--97166},
  year={2024}
}

@article{motiongpt,
  title={Motiongpt: Human motion as a foreign language},
  author={Jiang, Biao and Chen, Xin and Liu, Wen and Yu, Jingyi and Yu, Gang and Chen, Tao},
  journal={Advances in Neural Information Processing Systems},
  volume={36},
  pages={20067--20079},
  year={2023}
}

@article{shapegpt,
  title={Shapegpt: 3d shape generation with a unified multi-modal language model},
  author={Yin, Fukun and Chen, Xin and Zhang, Chi and Jiang, Biao and Zhao, Zibo and Liu, Wen and Yu, Gang and Chen, Tao},
  journal={IEEE Transactions on Multimedia},
  year={2025},
  publisher={IEEE}
}

@article{mar,
  title={Autoregressive image generation without vector quantization},
  author={Li, Tianhong and Tian, Yonglong and Li, He and Deng, Mingyang and He, Kaiming},
  journal={Advances in Neural Information Processing Systems},
  volume={37},
  pages={56424--56445},
  year={2024}
}

@article{janus,
  title={Janus: Decoupling visual encoding for unified multimodal understanding and generation},
  author={Wu, Chengyue and Chen, Xiaokang and Wu, Zhiyu and Ma, Yiyang and Liu, Xingchao and Pan, Zizheng and Liu, Wen and Xie, Zhenda and Yu, Xingkai and Ruan, Chong and others},
  journal={arXiv preprint arXiv:2410.13848},
  year={2024}
}

@article{emu3,
  title={Emu3: Next-token prediction is all you need},
  author={Wang, Xinlong and Zhang, Xiaosong and Luo, Zhengxiong and Sun, Quan and Cui, Yufeng and Wang, Jinsheng and Zhang, Fan and Wang, Yueze and Li, Zhen and Yu, Qiying and others},
  journal={arXiv preprint arXiv:2409.18869},
  year={2024}
}

@article{DINO,
  title={Dinov2: Learning robust visual features without supervision},
  author={Oquab, Maxime and Darcet, Timoth{\'e}e and Moutakanni, Th{\'e}o and Vo, Huy and Szafraniec, Marc and Khalidov, Vasil and Fernandez, Pierre and Haziza, Daniel and Massa, Francisco and El-Nouby, Alaaeldin and others},
  journal={arXiv preprint arXiv:2304.07193},
  year={2023}
}

@article{deng2024autoregressive,
  title={Autoregressive Video Generation without Vector Quantization},
  author={Deng, Haoge and Pan, Ting and Diao, Haiwen and Luo, Zhengxiong and Cui, Yufeng and Lu, Huchuan and Shan, Shiguang and Qi, Yonggang and Wang, Xinlong},
  journal={arXiv preprint arXiv:2412.14169},
  year={2024}
}

@article{zhang2025packing,
  title={Packing Input Frame Context in Next-Frame Prediction Models for Video Generation},
  author={Zhang, Lvmin and Agrawala, Maneesh},
  journal={arXiv preprint arXiv:2504.12626},
  year={2025}
}

@misc{magi1,
      title={MAGI-1: Autoregressive Video Generation at Scale},
      author={Sand-AI},
      year={2025},
      url={https://static.magi.world/static/files/MAGI_1.pdf},
}

@article{huang2025step,
  title={Step-video-ti2v technical report: A state-of-the-art text-driven image-to-video generation model},
  author={Huang, Haoyang and Ma, Guoqing and Duan, Nan and Chen, Xing and Wan, Changyi and Ming, Ranchen and Wang, Tianyu and Wang, Bo and Lu, Zhiying and Li, Aojie and others},
  journal={arXiv preprint arXiv:2503.11251},
  year={2025}
}

@article{kong2024hunyuanvideo,
  title={Hunyuanvideo: A systematic framework for large video generative models},
  author={Kong, Weijie and Tian, Qi and Zhang, Zijian and Min, Rox and Dai, Zuozhuo and Zhou, Jin and Xiong, Jiangfeng and Li, Xin and Wu, Bo and Zhang, Jianwei and others},
  journal={arXiv preprint arXiv:2412.03603},
  year={2024}
}

@article{wan2025,
      title={Wan: Open and Advanced Large-Scale Video Generative Models}, 
      author={Ang Wang and Baole Ai and Bin Wen and Chaojie Mao and Chen-Wei Xie and Di Chen and Feiwu Yu and Haiming Zhao and Jianxiao Yang and Jianyuan Zeng and Jiayu Wang and Jingfeng Zhang and Jingren Zhou and Jinkai Wang and Jixuan Chen and Kai Zhu and Kang Zhao and Keyu Yan and Lianghua Huang and Mengyang Feng and Ningyi Zhang and Pandeng Li and Pingyu Wu and Ruihang Chu and Ruili Feng and Shiwei Zhang and Siyang Sun and Tao Fang and Tianxing Wang and Tianyi Gui and Tingyu Weng and Tong Shen and Wei Lin and Wei Wang and Wei Wang and Wenmeng Zhou and Wente Wang and Wenting Shen and Wenyuan Yu and Xianzhong Shi and Xiaoming Huang and Xin Xu and Yan Kou and Yangyu Lv and Yifei Li and Yijing Liu and Yiming Wang and Yingya Zhang and Yitong Huang and Yong Li and You Wu and Yu Liu and Yulin Pan and Yun Zheng and Yuntao Hong and Yupeng Shi and Yutong Feng and Zeyinzi Jiang and Zhen Han and Zhi-Fan Wu and Ziyu Liu},
      journal = {arXiv preprint arXiv:2503.20314},
      year={2025}
}

@inproceedings{livesketch,
  title={Breathing life into sketches using text-to-video priors},
  author={Gal, Rinon and Vinker, Yael and Alaluf, Yuval and Bermano, Amit and Cohen-Or, Daniel and Shamir, Ariel and Chechik, Gal},
  booktitle={Proceedings of the IEEE/CVF Conference on Computer Vision and Pattern Recognition},
  pages={4325--4336},
  year={2024}
}

@article{claude,
  title={The Claude 3 Model Family: Opus},
  author={Anthropic, AI},
  journal={Sonnet, Haiku},
  year={2024}
}

@article{makeavideo,
  title={Make-a-video: Text-to-video generation without text-video data},
  author={Singer, Uriel and Polyak, Adam and Hayes, Thomas and Yin, Xi and An, Jie and Zhang, Songyang and Hu, Qiyuan and Yang, Harry and Ashual, Oron and Gafni, Oran and others},
  journal={arXiv preprint arXiv:2209.14792},
  year={2022}
}

@article{modelscope,
  title={Modelscope text-to-video technical report},
  author={Wang, Jiuniu and Yuan, Hangjie and Chen, Dayou and Zhang, Yingya and Wang, Xiang and Zhang, Shiwei},
  journal={arXiv preprint arXiv:2308.06571},
  year={2023}
}

@inproceedings{mikudance,
  title={Mikudance: Animating character art with mixed motion dynamics},
  author={Zhang, Jiaxu and Zeng, Xianfang and Chen, Xin and Zuo, Wei and Yu, Gang and Tu, Zhigang},
  booktitle={Proceedings of the IEEE/CVF International Conference on Computer Vision},
  pages={19689--19699},
  year={2025}
}

@inproceedings{psnr,
  title={Image quality metrics: PSNR vs. SSIM},
  author={Hore, Alain and Ziou, Djemel},
  booktitle={2010 20th international conference on pattern recognition},
  pages={2366--2369},
  year={2010},
  organization={IEEE}
}

@inproceedings{autoregressiveresi,
  title={Autoregressive image generation using residual quantization},
  author={Lee, Doyup and Kim, Chiheon and Kim, Saehoon and Cho, Minsu and Han, Wook-Shin},
  booktitle={Proceedings of the IEEE/CVF conference on computer vision and pattern recognition},
  pages={11523--11532},
  year={2022}
}

@article{xu2025withanyone,
  title={Withanyone: Towards controllable and id consistent image generation},
  author={Xu, Hengyuan and Cheng, Wei and Xing, Peng and Fang, Yixiao and Wu, Shuhan and Wang, Rui and Zeng, Xianfang and Jiang, Daxin and Yu, Gang and Ma, Xingjun and others},
  journal={arXiv preprint arXiv:2510.14975},
  year={2025}
}

@article{scalingvideo,
  title={Scaling autoregressive video models},
  author={Weissenborn, Dirk and T{\"a}ckstr{\"o}m, Oscar and Uszkoreit, Jakob},
  journal={arXiv preprint arXiv:1906.02634},
  year={2019}
}

@article{lumos,
  title={Lumos-1: On autoregressive video generation from a unified model perspective},
  author={Yuan, Hangjie and Chen, Weihua and Cen, Jun and Yu, Hu and Liang, Jingyun and Chang, Shuning and Lin, Zhihui and Feng, Tao and Liu, Pengwei and Xing, Jiazheng and others},
  journal={arXiv preprint arXiv:2507.08801},
  year={2025}
}

@inproceedings{svgbuilder,
  title={SVGBuilder: Component-Based Colored SVG Generation with Text-Guided Autoregressive Transformers},
  author={Chen, Zehao and Pan, Rong},
  booktitle={Proceedings of the AAAI Conference on Artificial Intelligence},
  volume={39},
  number={3},
  pages={2358--2366},
  year={2025}
}

@article{g3pt,
  title={G3pt: Unleash the power of autoregressive modeling in 3d generation via cross-scale querying transformer},
  author={Zhang, Jinzhi and Xiong, Feng and Xu, Mu},
  journal={arXiv preprint arXiv:2409.06322},
  year={2024}
}

@article{gr3,
  title={Gr-3 technical report},
  author={Cheang, Chilam and Chen, Sijin and Cui, Zhongren and Hu, Yingdong and Huang, Liqun and Kong, Tao and Li, Hang and Li, Yifeng and Liu, Yuxiao and Ma, Xiao and others},
  journal={arXiv preprint arXiv:2507.15493},
  year={2025}
}

@article{ida,
  title={Ida-vlm: Towards movie understanding via id-aware large vision-language model},
  author={Ji, Yatai and Zhang, Shilong and Wu, Jie and Sun, Peize and Chen, Weifeng and Xiao, Xuefeng and Yang, Sidi and Yang, Yujiu and Luo, Ping},
  journal={arXiv preprint arXiv:2407.07577},
  year={2024}
}

@article{luenhancing,
  title={Enhancing video transformers for action understanding with vlm-aided training},
  author={Lu, Hui and Jian, Hu and Poppe, Ronald and Salah, Albert Ali},
  journal={arXiv preprint arXiv:2403.16128},
  year={2024}
}

@article{sdvlm,
  title={SD-VLM: Spatial Measuring and Understanding with Depth-Encoded Vision-Language Models},
  author={Chen, Pingyi and Lou, Yujing and Cao, Shen and Guo, Jinhui and Fan, Lubin and Wu, Yue and Yang, Lin and Ma, Lizhuang and Ye, Jieping},
  journal={arXiv preprint arXiv:2509.17664},
  year={2025}
}

@inproceedings{tuneavideo,
  title={Tune-a-video: One-shot tuning of image diffusion models for text-to-video generation},
  author={Wu, Jay Zhangjie and Ge, Yixiao and Wang, Xintao and Lei, Stan Weixian and Gu, Yuchao and Shi, Yufei and Hsu, Wynne and Shan, Ying and Qie, Xiaohu and Shou, Mike Zheng},
  booktitle={Proceedings of the IEEE/CVF international conference on computer vision},
  pages={7623--7633},
  year={2023}
}

@article{show1,
  title={Show-1: Marrying pixel and latent diffusion models for text-to-video generation},
  author={Zhang, David Junhao and Wu, Jay Zhangjie and Liu, Jia-Wei and Zhao, Rui and Ran, Lingmin and Gu, Yuchao and Gao, Difei and Shou, Mike Zheng},
  journal={International Journal of Computer Vision},
  volume={133},
  number={4},
  pages={1879--1893},
  year={2025},
  publisher={Springer}
}

@article{videotetris,
  title={Videotetris: Towards compositional text-to-video generation},
  author={Tian, Ye and Yang, Ling and Yang, Haotian and Gao, Yuan and Deng, Yufan and Chen, Jingmin and Wang, Xintao and Yu, Zhaochen and Tao, Xin and Wan, Pengfei and others},
  journal={Advances in Neural Information Processing Systems},
  volume={37},
  pages={29489--29513},
  year={2024}
}
}


\end{document}